\documentclass[letterpaper, 10 pt, conference]{ieeeconf}  

\IEEEoverridecommandlockouts                              

\usepackage{graphicx}
\usepackage{amsmath,amsthm,amssymb,amsfonts}
\usepackage[colorlinks=true, allcolors=blue]{hyperref}
\usepackage{cleveref}
\usepackage{url}
\usepackage{caption}
\usepackage{subcaption}
\usepackage{makecell}
\usepackage{multicol}
\usepackage{multirow}
\usepackage{booktabs}
\usepackage{caption}
\usepackage{subcaption}
\usepackage[export]{adjustbox}
\usepackage{cuted}

\DeclareMathOperator{\atantwo}{atan2}

\title{\LARGE \bf
Imagining The Road Ahead: \\
Multi-Agent Trajectory Prediction via Differentiable Simulation
}

\author{
Adam \'Scibior$^{*1,2}$
Vasileios Lioutas$^{1,2}$
Daniele Reda$^{1,2}$
Peyman Bateni$^{1,2}$
Frank Wood$^{1,2,3}$
\thanks{$^*$Correspondence to: \texttt{adam.scibior@inverted.ai}}
\thanks{$^{1}$Inverted AI,
$^{2}$University of British Columbia,
$^{3}$Mila}
}

\begin{document}

\maketitle
\thispagestyle{empty}
\pagestyle{empty}


\begin{abstract}
We develop a deep generative model built on a fully differentiable simulator for multi-agent trajectory prediction. 
Agents are modeled with conditional recurrent variational neural networks (CVRNNs), which take as input an ego-centric birdview image representing the current state of the world and output an action, consisting of steering and acceleration, which is used to derive the subsequent agent state using a kinematic bicycle model. The full simulation state is then differentiably rendered for each agent, initiating the next time step. We achieve state-of-the-art results on the INTERACTION dataset, using standard neural architectures and a standard variational training objective, producing realistic multi-modal predictions without any ad-hoc diversity-inducing losses. We conduct ablation studies to examine individual components of the simulator, finding that both the kinematic bicycle model and the continuous feedback from the birdview image are crucial for achieving this level of performance. We name our model ITRA, for ``Imagining the Road Ahead''.
\end{abstract}

\section{Introduction}

Predicting where other vehicles and roadway users are going to be in the next few seconds is a critical capability for autonomous vehicles at level three and above \cite{noauthor_j3016b_nodate, boudette_despite_2019}. Such models are typically deployed on-board in autonomous vehicles to facilitate safe path planning. Crucially, achieving safety requires such predictions to be diverse and multi-modal, to account for all reasonable human behaviors, not just the most common ones. \cite{chai_multipath_2019, cui_multimodal_2019}.

Most of the existing approaches to trajectory prediction \cite{lee_desire_2017, sadeghian_sophie_2018, gupta_social_2018, zhao_multi-agent_2019, zhao_tnt_2020} treat it as a regression problem, encoding past states of the world to produce a distribution over trajectories for each agent in a relatively black-box manner, without accounting for future interactions. Several notable papers \cite{rhinehart_precog_2019, tang_multiple_2019, casas_implicit_2020, alahi_social_2016} explicitly model trajectory generation as a sequential decision-making process, allowing the agents to interact in the future. In this work, we take this approach a step further, embedding a complete, but simple and cheap, simulator into our predictive models. The simulator accounts not only for the positions of different agents, but also their orientations, sizes, kinematic constraints on their motion, and provides a perception mechanism that renders the world as a birdview RGB image. Our entire model, including the simulator, is end-to-end differentiable and runs entirely on a GPU, enabling efficient training by gradient descent.

\begin{figure*}[t]
    \centering
    \includegraphics[width=\textwidth]{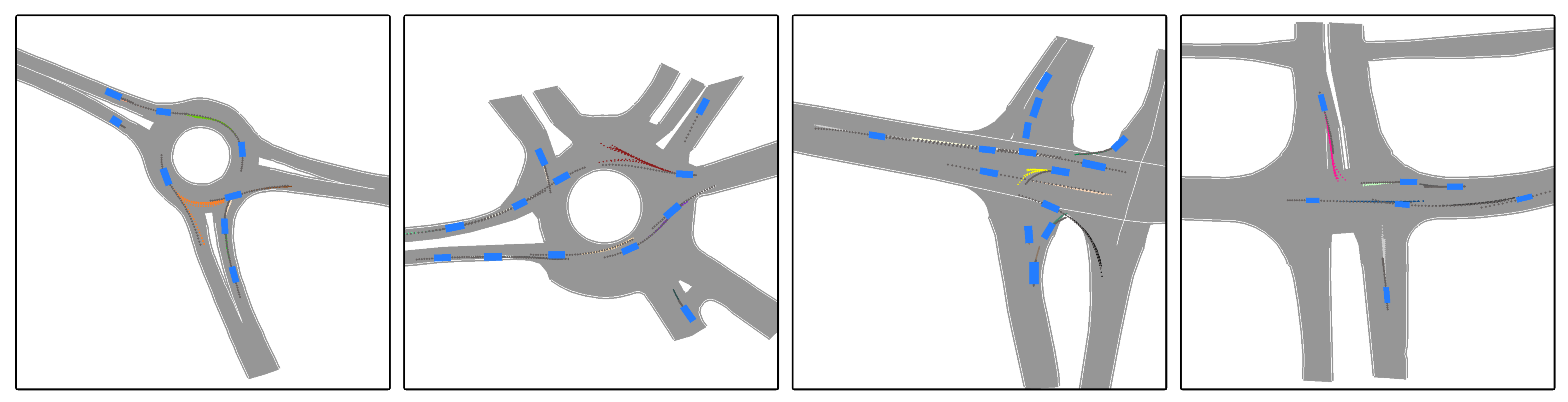}
    \captionof{figure}{Example predictions of ITRA for 3 seconds into the future based on a 1 second history. In most cases, the model is able to predict the ground truth trajectory with near certainty, but sometimes, such as when approaching a roundabout exit, there is inherent uncertainty which gives rise to multi-modal predictions. The ground truth trajectory is plotted in dark gray.}
    \label{fig:examples}
\end{figure*}

Each agent in our model is a separate conditional variational recurrent neural network (CVRNN), sharing network parameters across agents, which takes as input the birdview image provided by the simulator and outputs an action that is fed into the simulator. We use simple, well-established convolutional and recurrent neural architectures and train them entirely by maximum likelihood, using the tools of variational inference \cite{wainwright_graphical_2008, blei_variational_2017}. We do not rely on any ad-hoc featurization, problem-specific neural architectures, or diversity-inducing losses and still achieve new state-of-the-art performance and realistic multi-modal predictions, as shown in Figure \ref{fig:examples}.

We evaluate the resulting model, which we call ITRA for ``Imagining the Road Ahead'', on the challenging INTERACTION dataset \cite{zhan_interaction_2019} which itself is composed of highly interactive motions of road agents in a variety of driving scenarios in locations from different countries. ITRA outperforms all the published baselines, including the winners of the recently completed INTERPRET Challenge \cite{mo_recog_2020}.

\section{Background}

\subsection{Problem Formulation}

We assume that there are $N$ agents, where $N$ varies between test cases. At time $t \in 1, \dots, T$ each agent $i \in 1, \dots, N$ is described by a 4-dimensional state vector $s^i_t = (x^i_t, y^i_t, \psi^i_t, v^i_t)$, which includes the agents' position, orientation, and speed in a fixed global frame of reference. Each agent is additionally characterised by its immutable length $l^i$ and width $w^i$. We write $s_t = (s^1_t, \dots, s^N_t)$ for the joint state of all agents at time $t$. The goal is to predict future trajectories $s^i_t$ for $t \in T_{obs}+1:T$ and $i \in 1:N$, based on past states $1:T_{obs}$. This prediction task can technically can be framed as a regression problem, and as such, in principle, could be solved by any regression algorithm.  We find, however, that it is beneficial to exploit the natural factorization of the problem in time and across agents, and to incorporate some domain knowledge in a form of a simulator.

\subsection{Kinematic Bicycle Model}

\begin{figure}[t]
    \centering
    \includegraphics[width=\linewidth]{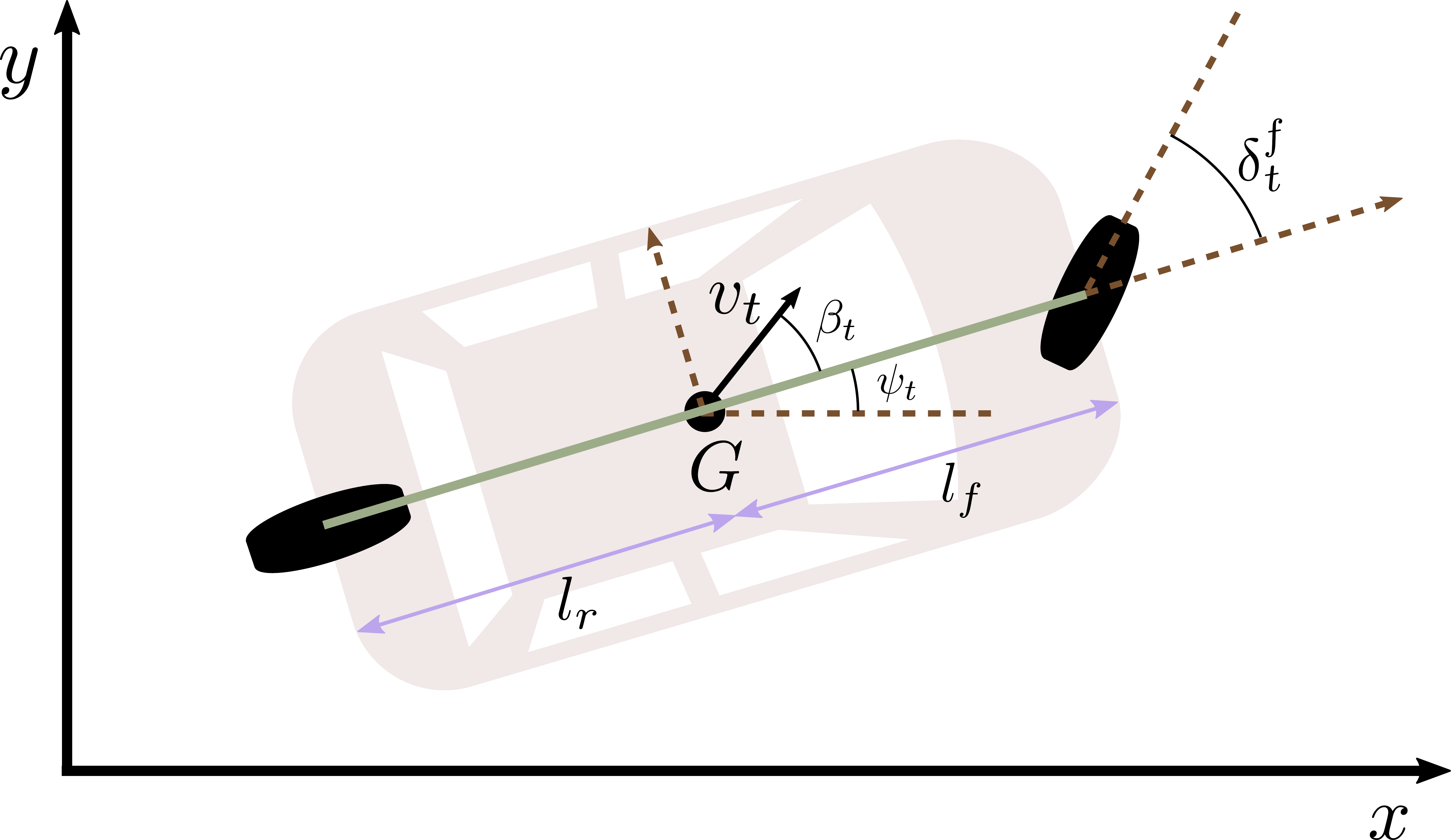}
    \caption{\cite{polack_kinematic_2017} The kinematic bicycle model. $G$ is the geometric center of the vehicle and $v_t$ is its instantaneous velocity. Since our dataset does not contain entries for $\delta_t^f$, we regard $\beta_t$ as steering directly, which is mathematically equivalent to setting $l_f = 0$.
    We fit $l_r$ by grid search for each recorded vehicle trajectory.} 
    \label{fig:bicycle-model}
\end{figure}

The bicycle kinematic model \cite{rajamani_lateral_2012}, depicted in Figure \ref{fig:bicycle-model}, is known to be an accurate model of vehicle motion when not skidding or sliding and we found it to near-perfectly describe the trajectories of vehicles in the INTERACTION dataset. The action space in the bicycle model consists of steering and acceleration.
We provide detailed equations of motion and discuss the procedure for fitting the bicycle model in the Appendix.

\subsection{Variational Autoencoders}

Variational Autoencoders (VAEs) \cite{kingma_auto-encoding_2014} are a very popular class of deep generative models, which consist of a simple latent variable and a neural network which transforms its value into the parameters of a simple likelihood function. The neural network parameters are optimized by gradient ascent on the variational objective called the evidence lower bound (ELBO), which approximates maximum likelihood learning. To extend this model to a sequential setting, we employ variational recurrent neural networks (VRNNs) \cite{chung_recurrent_2016}. In our setting, we have additional context provided as input to those models, so they are technically  \emph{conditional}, that is CVAEs and CVRNNs respectively.

\section{Method}

\begin{figure*}
    \centering
    \includegraphics[width=0.9\textwidth]{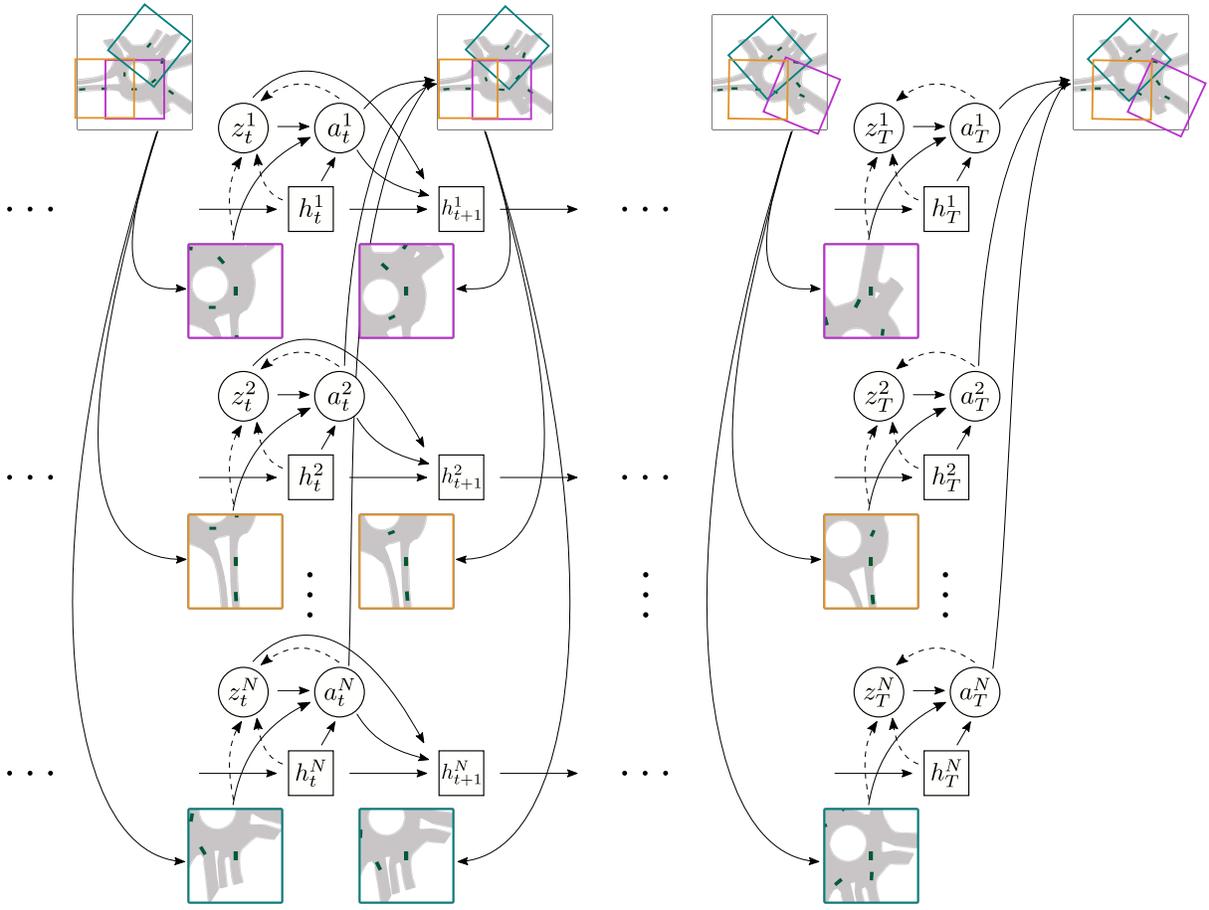}
    \caption{A schematic illustration of ITRA's architecture. The agents choose the actions at each time step independently from each other, obtaining information about other agents and the environment only through ego-centered birdview images. This representation naturally accommodates an unbounded, variable number of agents and time steps. The actions are translated into new agent states using a kinematic model not shown on the diagram. The colored bounding boxes illustrate each agent's field of view in the full scene.}
    \label{fig:model}
\end{figure*}

Our model consists of a map specifying the driveable area and a composition of homogeneous agents, whose interactions are mediated by a differentiable simulator. At each timestep $t$, each agent $i$ chooses an action $a^i_t$, which then determines its next state according to a kinematic model $s^i_{t+1} = \mathop{kin}(s^i_t, a^i_t)$, which is usually the bicycle model, but we also explore some alternatives in Section \ref{sec:kinematic}. The states of all agents are pooled together to form the full state of the world $s_t$, which is presented to each agent as $b^i_t$, a rasterized, ego-centered and ego-rotated birdview image, as depicted in Figure \ref{fig:model}. This rasterized birdview representation is commonly employed in existing models \cite{djuric_uncertainty-aware_2020, zhao_multi-agent_2019, tang_multiple_2019}. We choose to use three RGB channels to reduce the embedding network size. Each agent has its own hidden state $h^i_t$, which serves as the agent's memory, and a source of randomness $z^i_t$, which is a random variable that is used to generate the agent's stochastic behavior. The full generative model then factorizes as follows
\begin{align}
    &p(s_{1:T}) = \\ &\int \int \prod_{t=1}^T \prod_{i=1}^N p(z^i_t) p(a^i_t | b^i_t, z^i_t, h^i_t) p(s^i_{t+1} | s^i_t, a^i_t) \ dz^{1:N}_{1:T} \ da^{1:N}_{1:T} \nonumber
\end{align}
Our model uses additional information about the environment in a form of a map rendered onto the birdview image, but for clarity we omit this from our notation.

In ITRA, the conditional distribution of actions given latent variables and observations $p(a^i_t | b^i_t, z^i_t, h^i_t)$ is deterministic, the distribution of the latent variables is unit Gaussian $p(z^i_t) = \mathcal{N}(0,1)$, and the distribution of states is Gaussian with a mean dictated by the kinematic model and a fixed diagonal variance $p(s^i_{t+1} | s^i_t, a^i_t) = \mathcal{N}(\mathop{kin}(s^i_t, a^i_t), \sigma \bf{I})$, where $\sigma$ is a hyperparameter of the model.

\subsection{Differentiable Simulation}

While the essential modeling of human behavior is performed by the VRNN generating actions, an integral part of ITRA is embedding that VRNN in a simulation consisting of the kinematic model and the procedure for generating birdview images, which is used both at training and test time. The entire model is end-to-end differentiable, allowing the parameters of the VRNN to be trained by gradient descent. For the kinematic model, this is as simple as implementing its equations of motion in PyTorch \cite{paszke_pytorch_2019}, but the birdview image construction is more involved.

We use PyTorch3D \cite{ravi_accelerating_2020}, a PyTorch-based library for differentiable rendering, to construct the birdview images. While the library primarily targets 3D applications, we obtain a simple 2D rasterization by applying an orthographic projection camera and a minimal shader that ignores lighting. To make the process differentiable we use the soft RGB blend algorithm \cite{liu_soft_2019}.

\subsection{Neural Networks}

Each agent is modelled with a two-layer recurrent neural network using the gated recurrent unit (GRU) with a convolutional neural network (CNN) encoder for processing the birdview image. The remaining components are fully connected. We choose the latent dimensions to be $64$ for $h$ and $2$ for $z$ and we use birdview images in $256{\times}256$ resolution corresponding to a $100{\times}100$ meter area. We chose standard, well-established architectures to demonstrate that the improved performance in ITRA comes from the inclusion of a differentiable simulator.

\subsection{Training and Losses}

We train all the network components jointly from scratch using the evidence lower bound (ELBO) as the optimization objective. For this purpose, we use an inference network that parameterizes a diagonal Gaussian distribution over $z$ given the current observation and recurrent state, as well as the ground truth action extracted from the data, defining a variational distribution $q(z^i_t | a^i_t, b^i_t, h^i_t)$. The ELBO is then defined as
\begin{align}
\label{eq:elbo}
\begin{split}
    \mathcal{L} = \sum_{i=1}^{N} \sum_{t=1}^{T-1} \Big( & \mathbb{E}_{q(z^i_t | a^i_t, b^i_t, h^i_t)} \left[ \log p(s^i_{t+1} | b^i_t, z^i_t, h^i_t) \right] \\ 
    & - KL\left[q(z^i_t | a^i_t, b^i_t, h^i_t) || p(z^i_t) \right] \Big),
\end{split}
\end{align}
where $s^i_t$ are the ground truth states obtained from the dataset. This is the standard setting for training conditional variational recurrent neural networks (CVRNNs), with a caveat that we introduce a distinction between actions and states.

\subsection{Conditional Predictions}

Typically, trajectory prediction tasks are defined as predicting a future trajectory $T_{obs}+1:T$ based on a past trajectory $1:T_{obs}$. Predicting trajectories from a single frame is unnecessarily difficult and using Eq.~\ref{eq:elbo} naively with a randomly initialized recurrent state $h_1$ destabilizes the training process and leads to poor predictions. To make sure the recurrent state is seeded well, we employ teacher forcing for states $s_t$ at time $1:T_{obs}$ but not after $T_{obs}$, which corresponds to performing an inference over the state of $h_{T_{obs}}$ given observations from previous time steps. We do this both at training and test time.

\section{Experimental Results}

For our experiments, we use the INTERACTION dataset \cite{zhan_interaction_2019}, which consists of about 10 hours of traffic data containing 36,000 vehicles recorded in 11 locations around the world at intersections, roundabouts, and highway ramps. Our model achieves state-of-the-art performance, as reported in Table \ref{tab:main-results}, and we also provide in Table \ref{tab:individual-scenes} a more detailed breakdown of the scores it achieves across different scenes to enable a more fine-grained analysis. Finally, we ablate the kinematic bicycle model and the birdview generation for future times, the two key components of ITRA's simulation, showing that both of them are necessary to achieve the reported performance.

Following the recommendation of \cite{tang_multiple_2019}, we apply classmates forcing, where the actions and states of all agents other than the ego vehicle are fixed to ground truth at training time and not generated by the VRNN, which allows us to use batches of diverse examples and stabilizes training. At test time, we predict the motion of all agents in the scene, including bicycles/pedestrians (which are not distinguished from each other in the dataset). For simplicity, we use the same model trained on vehicles to predict all agent types, noting that it could be possible to obtain further improvements by training a separate model for bicycles/pedestrians. We trained all the model components jointly from scratch using the ADAM optimizer \cite{kingma_adam_2017} with the standard learning rate of 3e-4, using gradient clipping and a batch size of 8. We found the training to be relatively stable with respect to other hyperparameters and have not performed extensive tuning. The training process takes about two weeks on a single NVIDIA Tesla P100 GPU.

Finally, we note that we compute ground truth actions used as input to the inference network based on a sequence of ground truth states, independently of what the model predicts at earlier times. We found that this greatly speeds up training and leads to better final performance.

\begin{table}[t]
    \centering
    \caption{Validation set prediction errors on the INTERACTION Dataset evaluated with the suggested train/validation split. The evaluation is based on six samples, where for each example the trajectory with the smallest error is selected, independently for ADE and FDE. ReCoG only makes deterministic predictions, so its performance would be the same with a single sample. The values for baselines are as reported by their original authors, with the exception of DESIRE and MultiPath, where we present the results reported in \cite{zhao_tnt_2020}.}
    \label{tab:main-results}
    \begin{tabular}{l|cc}
    \hline
    Method & $\min\text{ADE}_6$ & $\min\text{FDE}_6$ \\
    \hline
    DESIRE\cite{lee_desire_2017} & 0.32 & 0.88 \\
    MultiPath\cite{chai_multipath_2019} & 0.30 & 0.99 \\
    TNT\cite{zhao_tnt_2020} & 0.21 & 0.67 \\
    ReCoG\cite{mo_recog_2020} & 0.19 & 0.65 \\
    ITRA (ours) & \bf{0.17} & \bf{0.49} \\
    \hline
    \end{tabular}
\end{table}

\begin{table}[]
\centering
\caption{Breakdown of ITRA's performance across individual scenes in the INTERACTION dataset. We present evaluation of a single model trained on all scenes jointly, noting that fine-tuning on individual scenes can further improve performance. CHN Merging ZS, which is a congested highway where predictions are relatively easy, is overrepresented in the dataset, lowering the average error across scenes.}
\label{tab:individual-scenes}
\begin{tabular}{l|ccc}
\toprule
Scene & $\min\text{ADE}_6$ & $\min\text{FDE}_6$ & $\text{MFD}_6$\\
\midrule
CHN Merging ZS & 0.127 & 0.356 & 2.157 \\
CHN Roundabout LN & 0.199 & 0.549 & 2.354 \\
DEU Merging MT & 0.226 & 0.678 & 2.178 \\
DEU Roundabout OF & 0.283 & 0.766 & 3.389 \\
USA Intersection EP0 & 0.215 & 0.640 & 3.008 \\
USA Intersection EP1 & 0.234 & 0.678 & 2.974 \\
USA Intersection GL & 0.203 & 0.611 & 2.750 \\
USA Intersection MA & 0.212 & 0.634 & 3.430 \\
USA Roundabout EP & 0.234 & 0.690 & 3.094 \\
USA Roundabout FT & 0.219 & 0.661 & 2.932 \\
USA Roundabout SR & 0.178 & 0.525 & 2.451 \\
\bottomrule
\end{tabular}
\end{table}

\begin{figure*}[t]
    \centering
    \begin{subfigure}[b]{0.49\columnwidth}
        \centering
        \includegraphics[width=\columnwidth,frame]{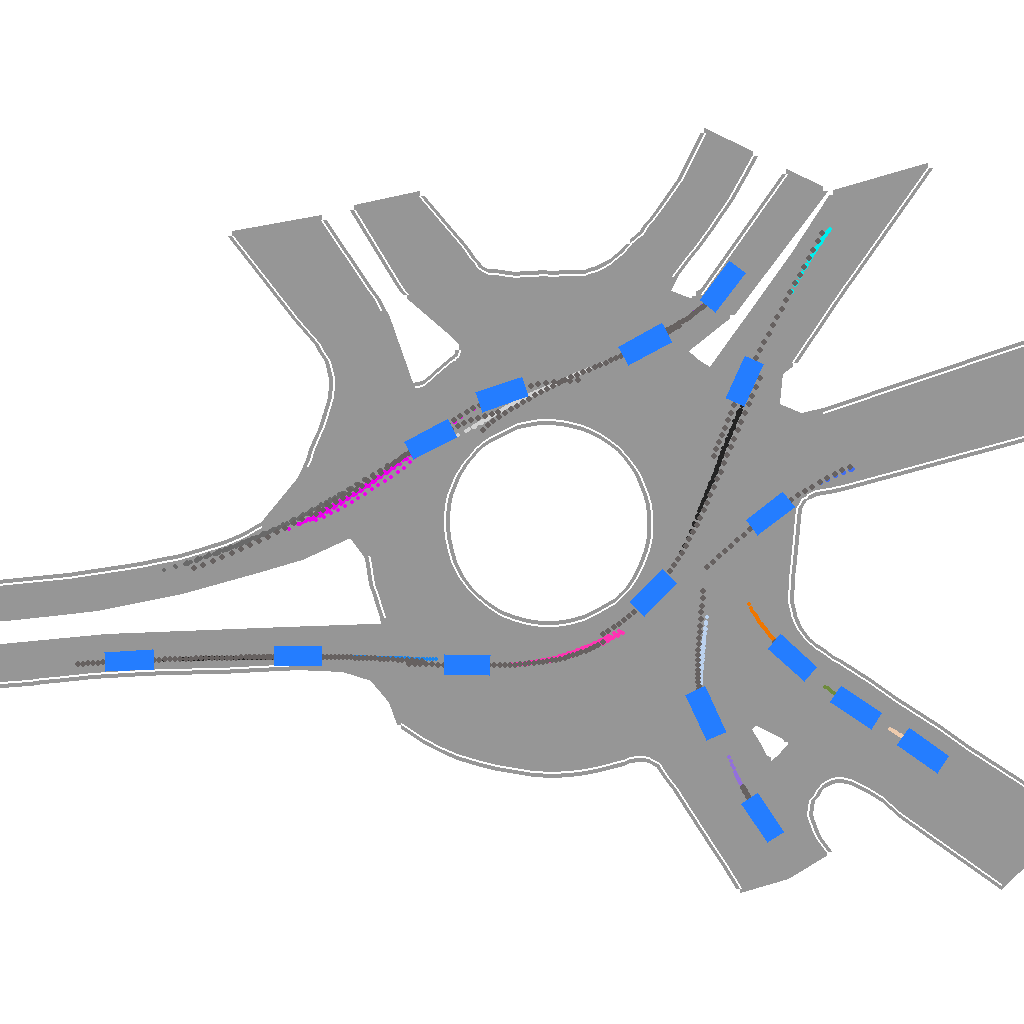}
        \caption{$t = 0s$}
        \label{fig:multi_pred_a}
    \end{subfigure}
    \begin{subfigure}[b]{0.49\columnwidth}
        \centering
        \includegraphics[width=\columnwidth, frame]{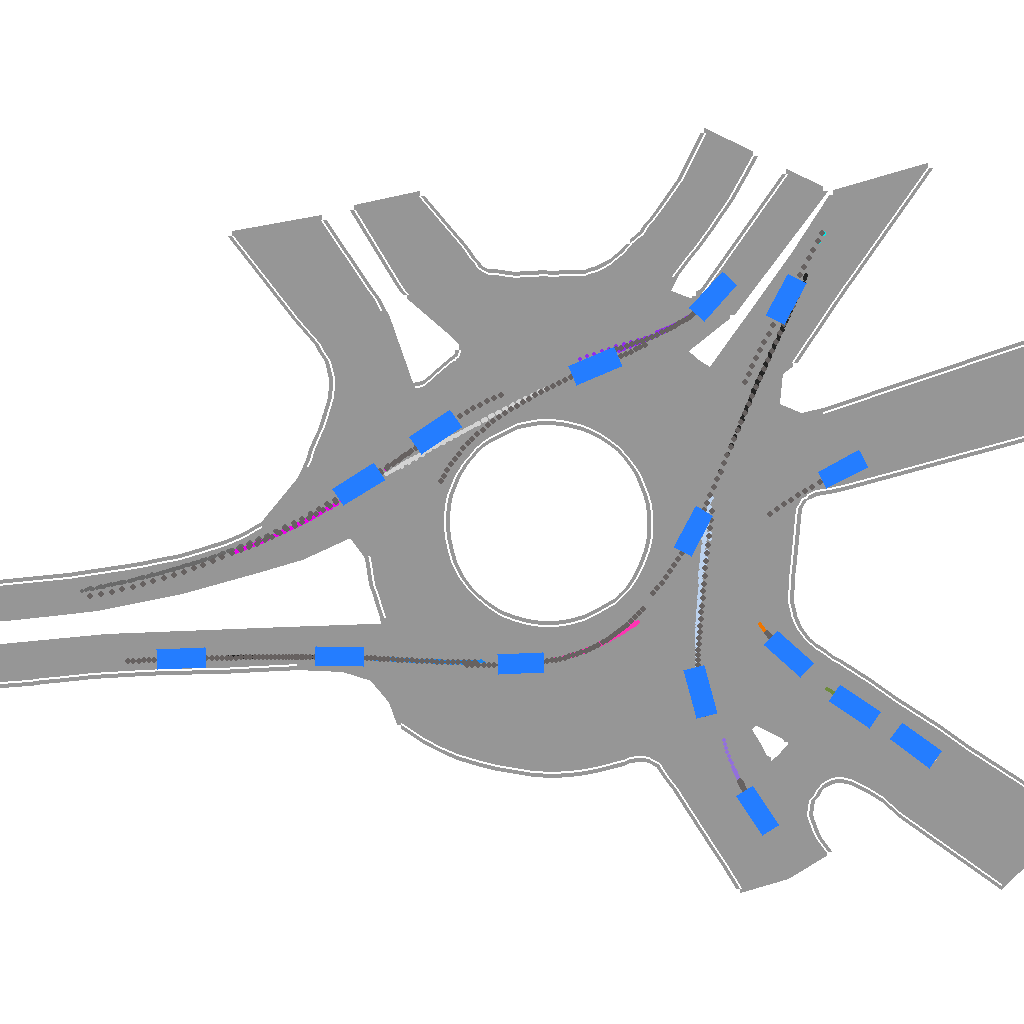}
        \caption{$t = 1s$}
        \label{fig:multi_pred_b}
    \end{subfigure}
    \begin{subfigure}[b]{0.49\columnwidth}
        \centering
        \includegraphics[width=\columnwidth, frame]{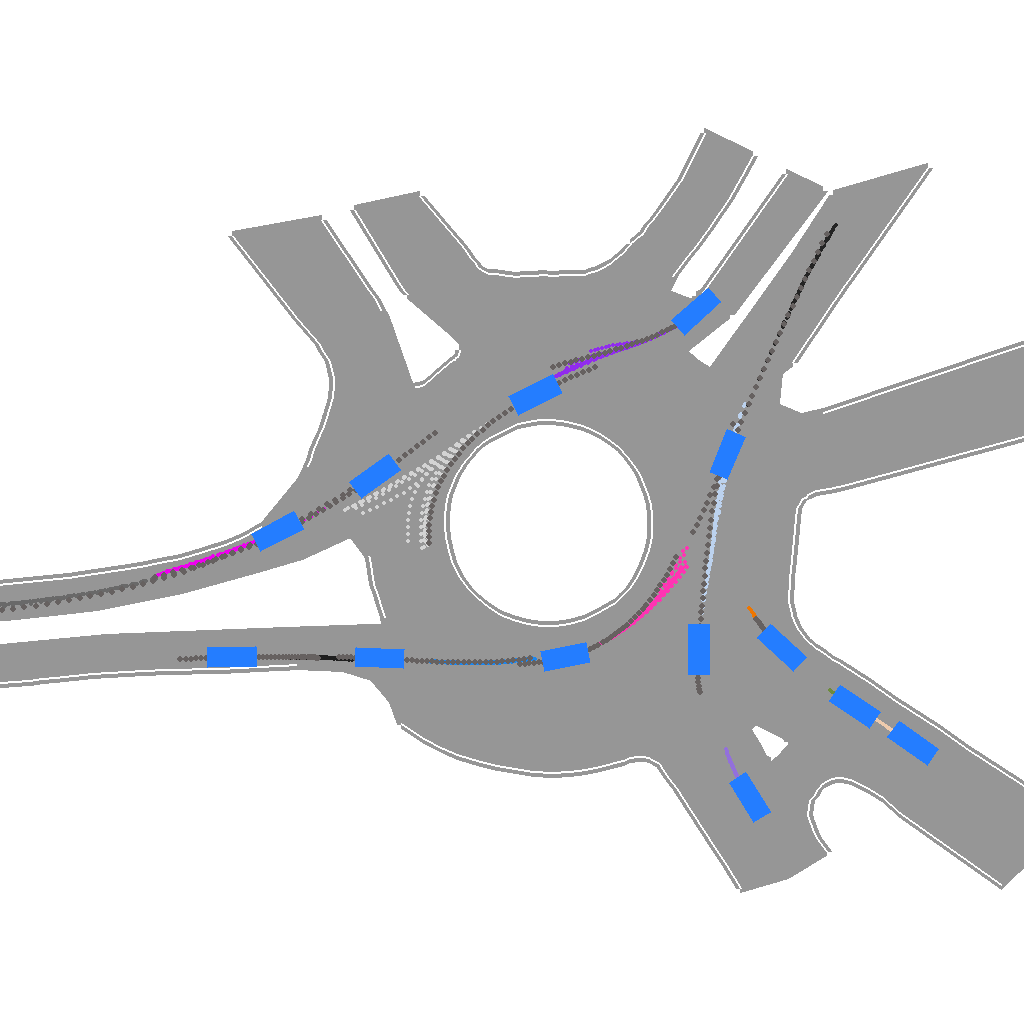}
        \caption{$t = 2s$}
        \label{fig:multi_pred_c}
    \end{subfigure}
    \begin{subfigure}[b]{0.49\columnwidth}
        \centering
        \includegraphics[width=\columnwidth, frame]{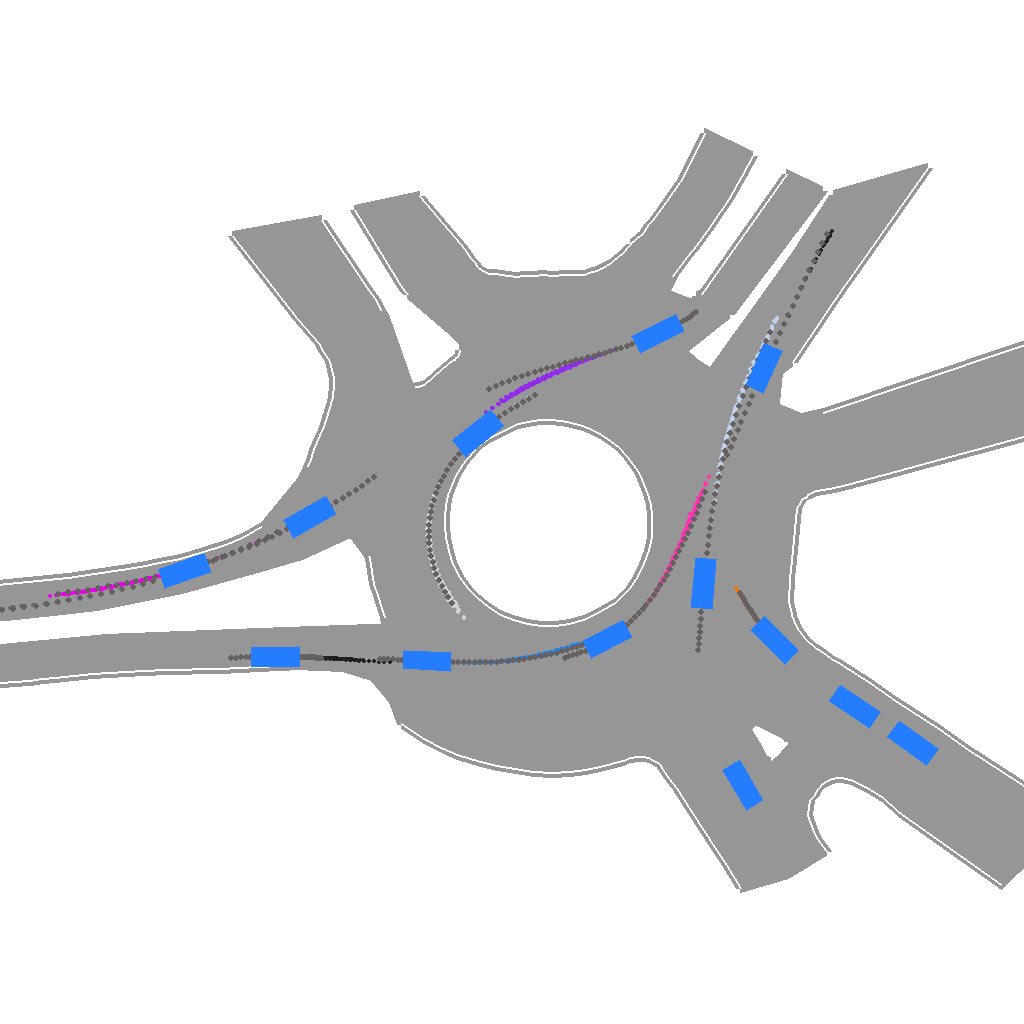}
        \caption{$t = 3s$}
        \label{fig:multi_pred_d}
    \end{subfigure}
    \caption{A sequence of 4 subsequent prediction snapshots spaced 1 second apart. For each vehicle in the scene we show 10 sampled predictions over the subsequent 3 seconds based on the previous 1 second. Notice the car that just entered the roundabout from top-right in \ref{fig:multi_pred_a}, for which ITRA predicts that it will go forward in the roundabout, and as it enters the roundabout in \ref{fig:multi_pred_c}, the prediction diversifies to include both exiting the roundabout and continuing the turn, finally matching the ground truth trajectory (in dark grey) in \ref{fig:multi_pred_d} once the car started turning.}
    \label{fig:multi_prediction}
\end{figure*}

\subsection{Evaluation Metrics} \label{sec:eval-metrics}

Typical multi-agent predictive model evaluation allows up to $K$ predicted trajectories to be generated for each agent, that is $(x^{i,k}_t, y^{i,k}_t)$ for $i \in 1:N, t \in T_{obs}+1:T, k \in 1:K$. Models are compared via the average displacement error (ADE) and the final displacement error (FDE), defined as
\begin{align*}
    \text{ADE}^i_k &= \sqrt{\frac{1}{(T - T_{obs})} \sum_{t=T_{obs}+1}^T (x^{i,k}_t - x^i_t) ^ 2 + (y^{i,k}_t - y^i_t) ^ 2} \\
    \text{FDE}^i_k &= \sqrt{(x^{i,k}_T - x^i_T) ^ 2 + (y^{i,k}_T - y^i_T) ^ 2} ,
\end{align*}
where $(x^i_t, y^i_t)$ is the ground truth position of agent $i$ at time $t$. When multiple trajectory predictions can be generated, the predicted trajectory with the smallest error is selected for evaluation. In either case, the errors are averaged across agents and test examples.
\begin{align*}
    \min\text{ADE}_K &= \frac{1}{N} \sum_{i=1}^N \mathop{\min}_k \text{ADE}^i_k \\
    \min\text{FDE}_K &= \frac{1}{N} \sum_{i=1}^N \mathop{\min}_k \text{FDE}^i_k
\end{align*}
Note that these metrics only score position and not orientation or speed predictions, so most trajectory prediction models only generate position coordinates.

While not directly measuring performance, a useful metric for tracking the diversity of predicted trajectories is the Maximum Final Distance (MFD)
\begin{align*}
    \text{MFD}_K &= \frac{1}{N} \sum_{i=1}^N \mathop{\max}_{k, l} \sqrt{(x^{i,k}_T - x^{i,l}_T) ^ 2 + (y^{i,k}_T - y^{i,l}_T) ^ 2} ,
\end{align*}
which we primarily use to diagnose when the model collapses onto deterministic predictions.

\subsection{Ablations on Future Birdview Images}

A key aspect of ITRA is the ongoing simulation throughout the future time steps, generating updated birdview images providing feedback to the agents. While powerful, this simulation complicates the software architecture and is computationally expensive. In this section, we analyze two ablations, which remove the dynamic feedback aspect provided by future birdview images.

\begin{figure}
    \centering
    \includegraphics[width=\columnwidth]{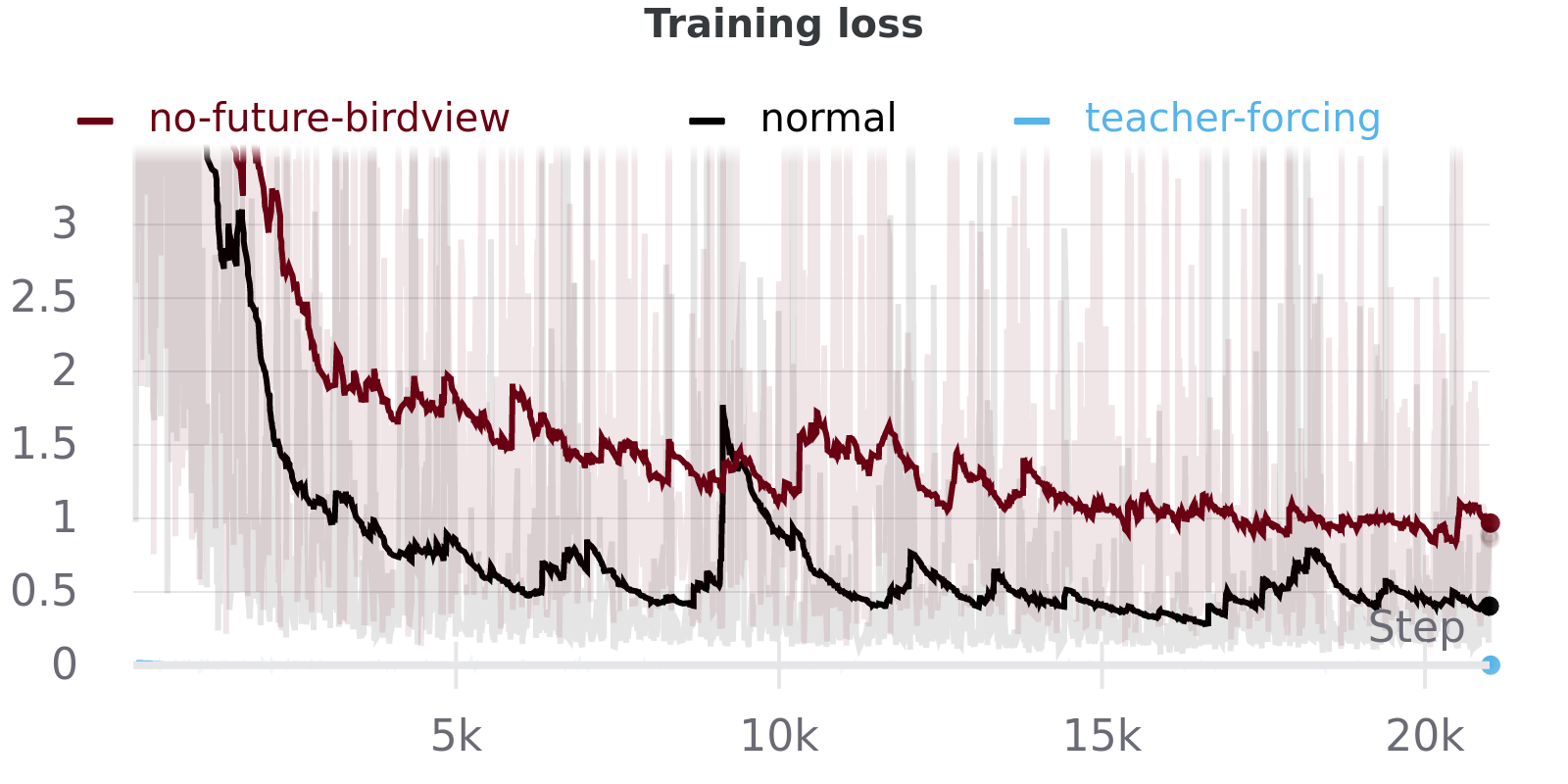}
    \includegraphics[width=\columnwidth]{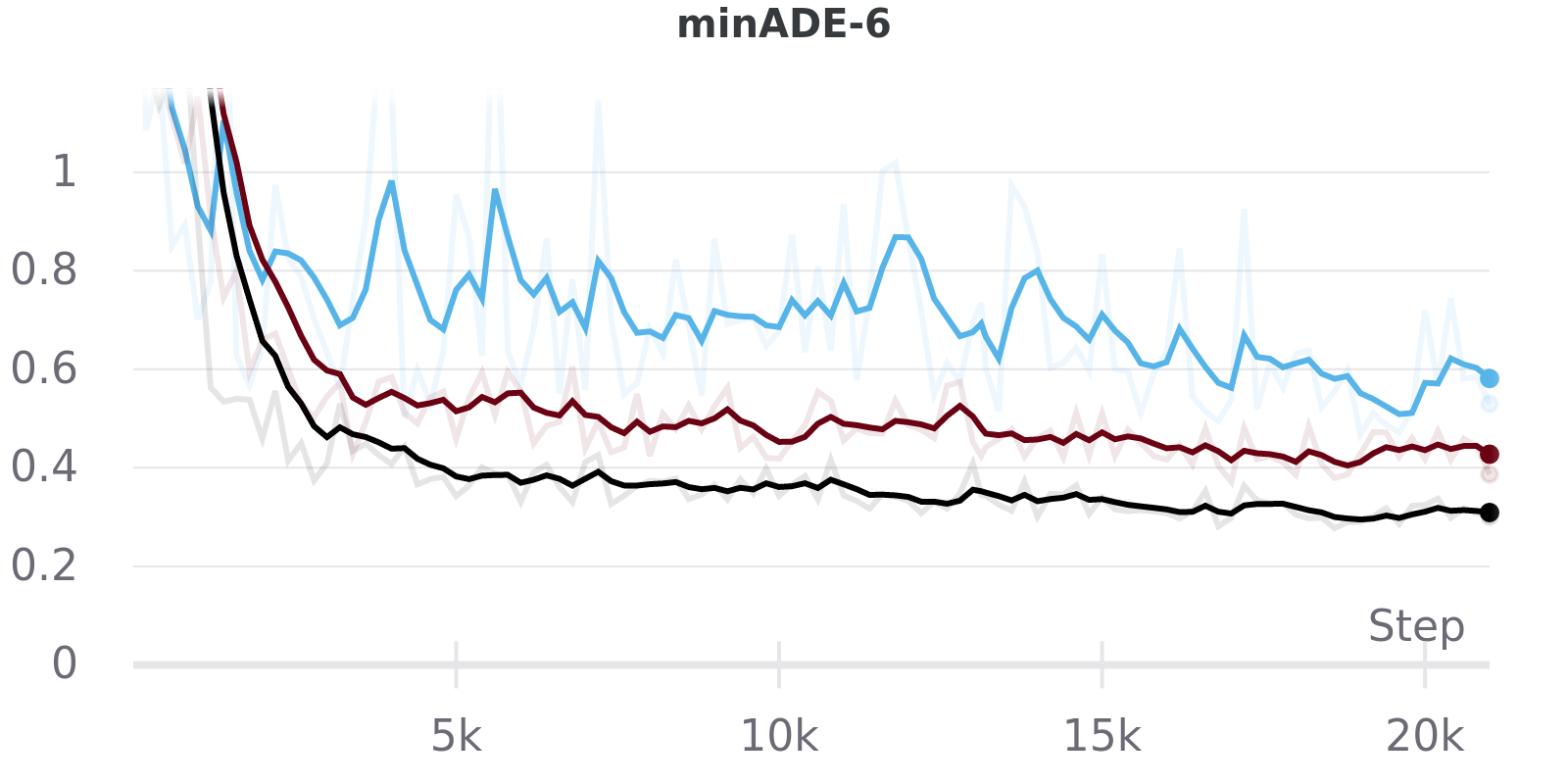}
    \includegraphics[width=\columnwidth]{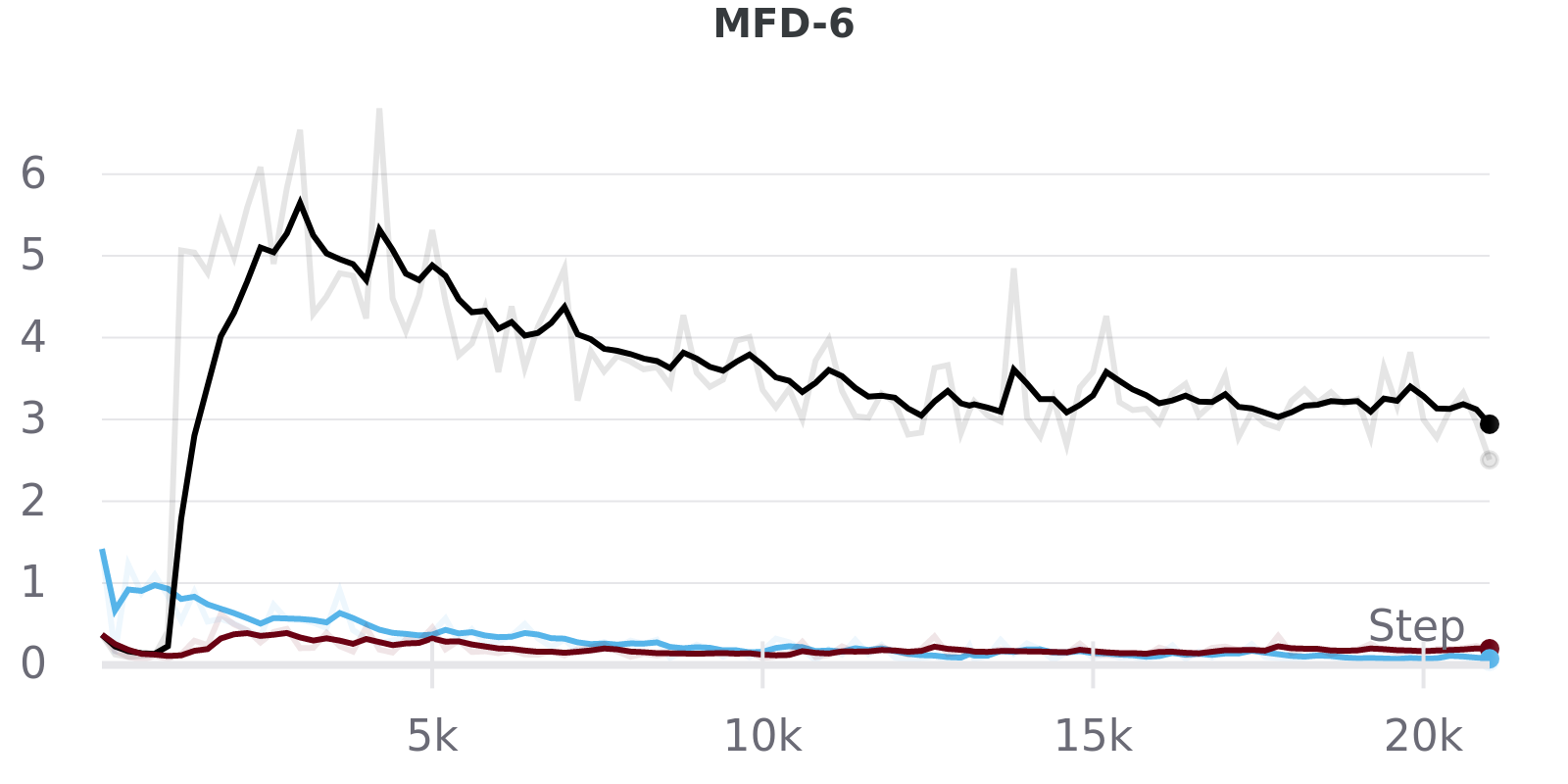}
    \caption{Comparison of a normal ITRA training process with one that uses a blank birdview image for future steps and with one that uses teacher forcing, i.e. uses birdview images representing ground truth at training time. Both of these ablations produce inaccurate results and essentially no diversity in predictions, indicating that it is necessary to generate birdview images dynamically both at training and test time. With teacher forcing the training loss is so small that it blends in with the x axis.}
    \label{fig:ablations}
\end{figure}

In the first ablation, we use normal birdview images up to time $T_{obs}$ and then blank images subsequently, both at training and test time. This removes any feedback and roughly corresponds to models that predict the entire future trajectory in a single step based on past observations \cite{lee_desire_2017, sadeghian_sophie_2018, gupta_social_2018, zhao_multi-agent_2019, zhao_tnt_2020}, albeit with a suboptimal neural architecture. In the second ablation, we perform teacher forcing at training time, fixing the agent's state and the birdview image to the ground truth values at each time step. In both cases, we find that the ablated model performs significantly worse and exhibits no prediction diversity, as shown in Figure \ref{fig:ablations}.

\subsection{Alternative Kinematic Models} \label{sec:kinematic}

\begin{figure}
    \centering
    \includegraphics[width=\columnwidth]{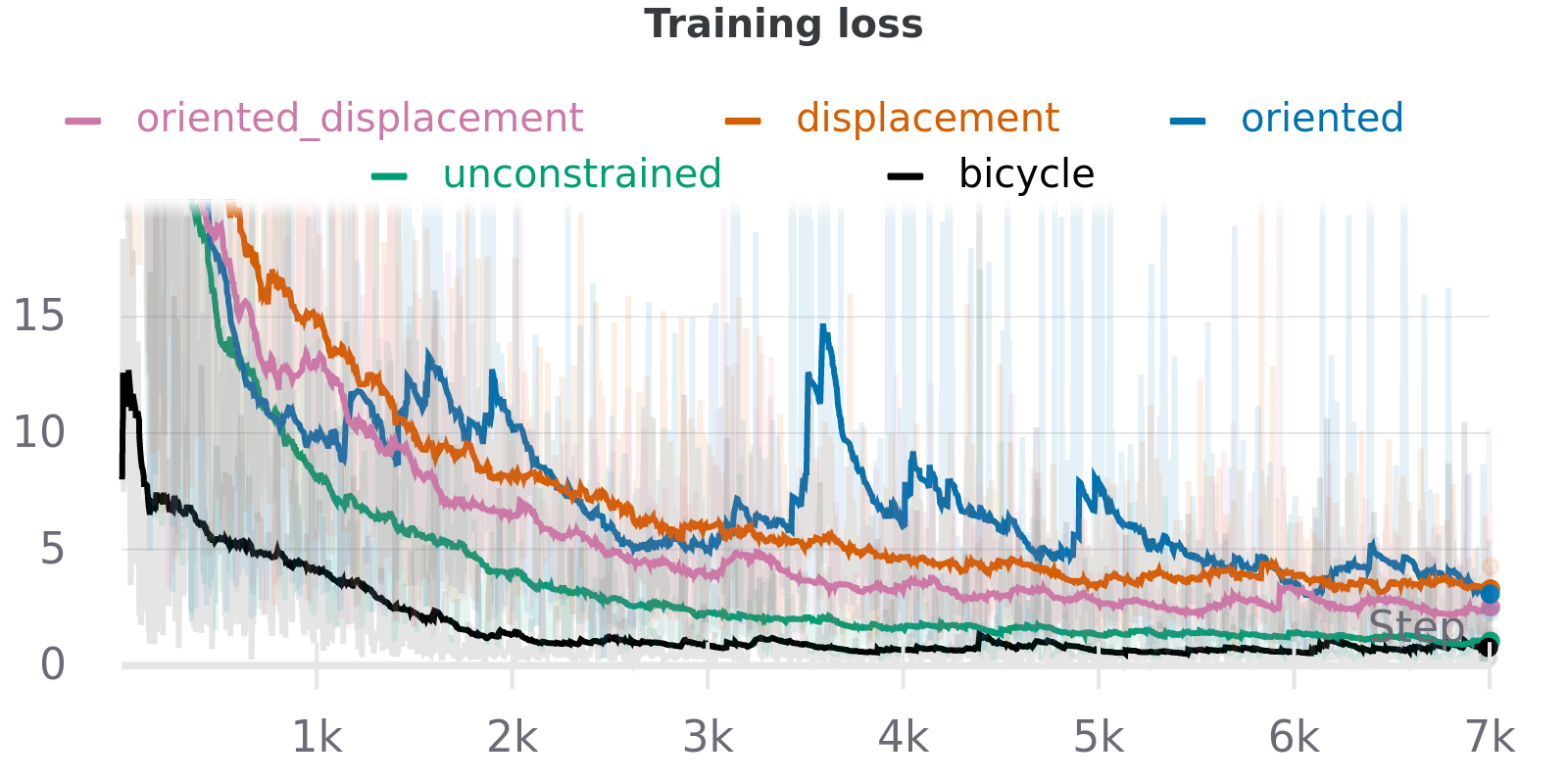}
    \includegraphics[width=\columnwidth]{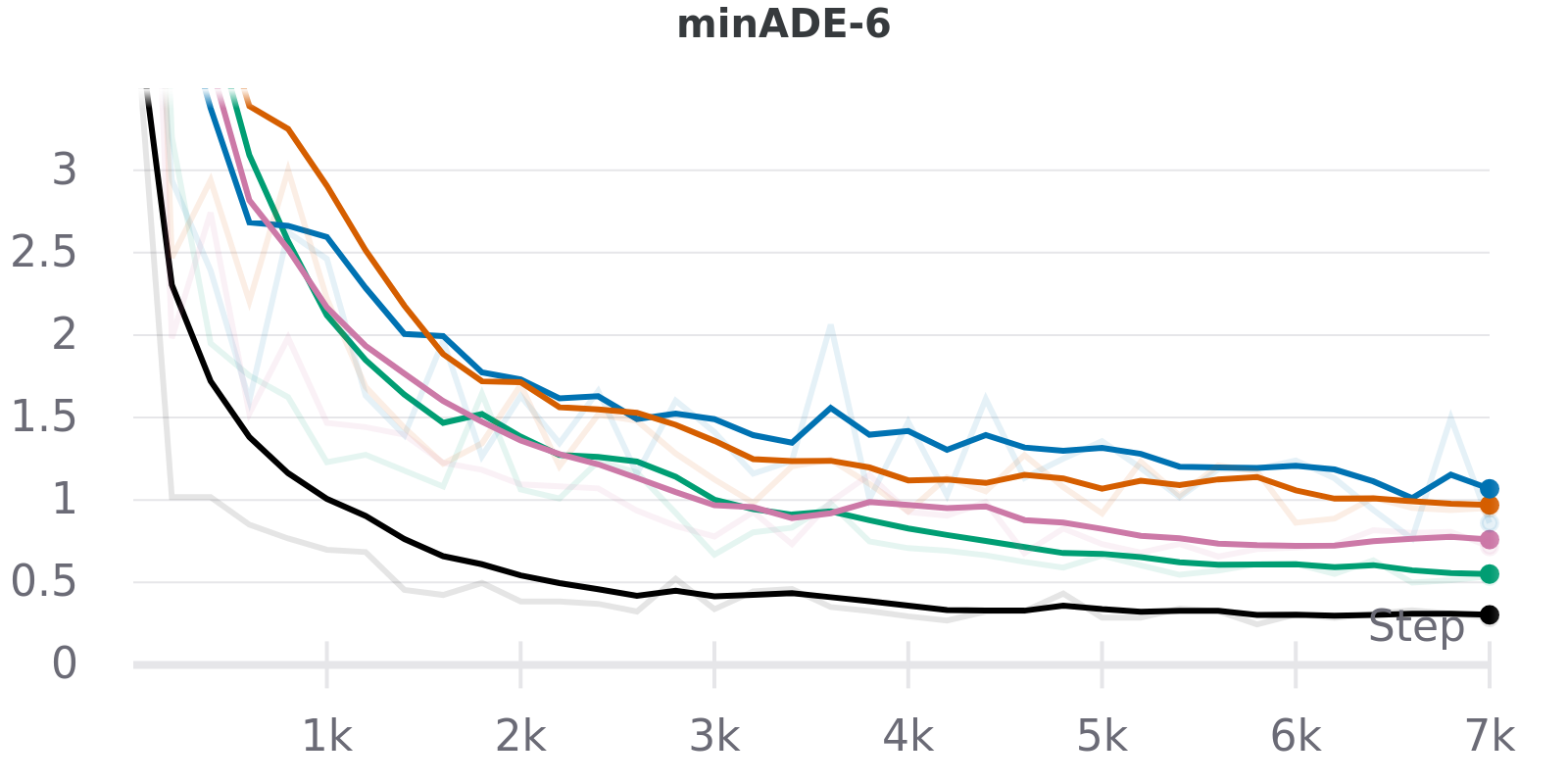}
    \includegraphics[width=\columnwidth]{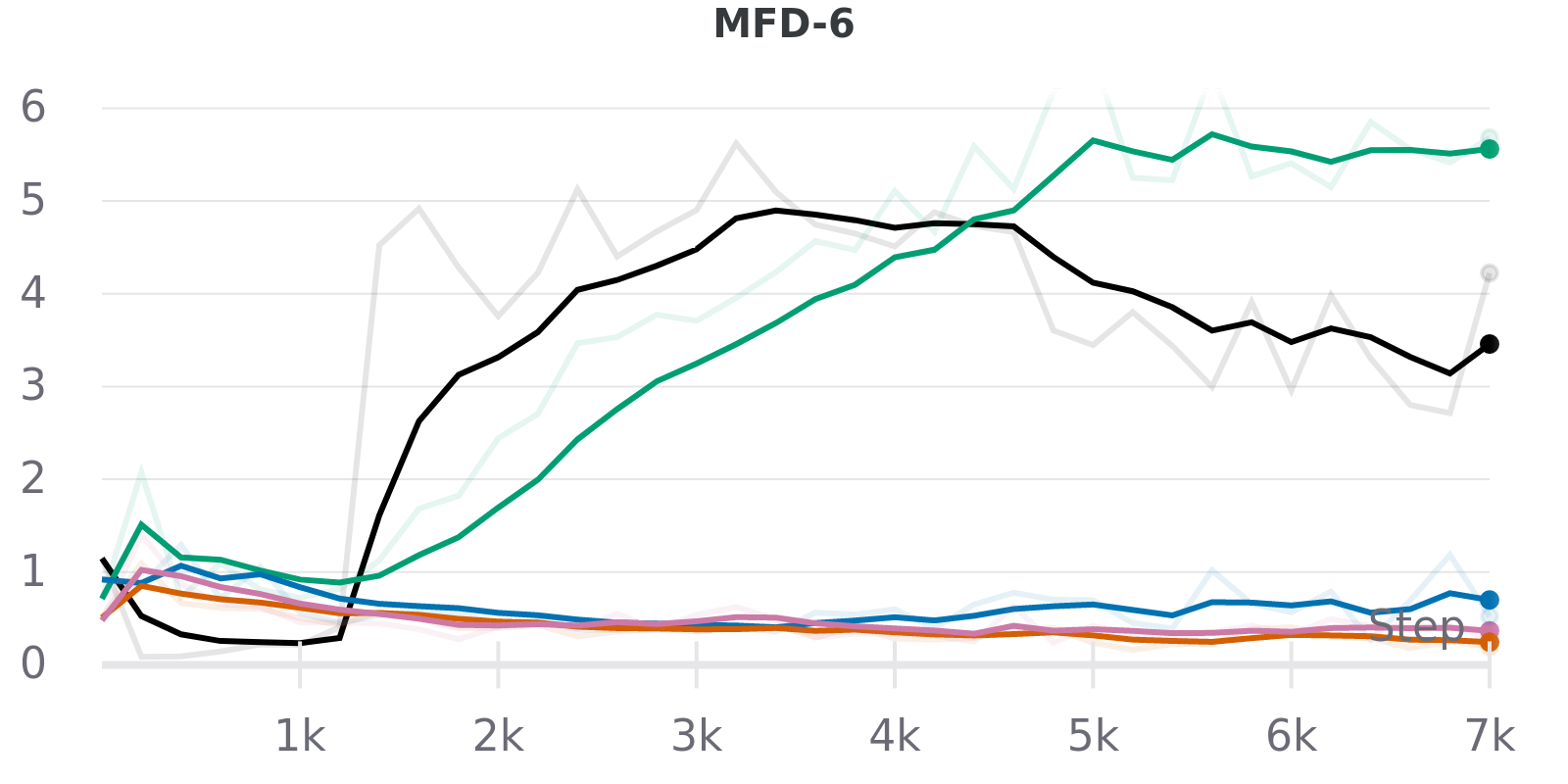}
    \caption{Training curves for ITRA with different kinematic models. The bicycle model with its standard action space outperforms all the alternatives by a substantial margin. We smoothed the curves using exponential moving average to make the plots more legible, leaving the raw measurements faded out.}
    \label{fig:kinematic-ablation}
\end{figure}

While our approach relies on the kinematic bicycle model, we have also investigated several alternatives. In the simplest one, which we call the \emph{unconstrained model}, the action is the delta between subsequent state vectors.
\begin{align}
    a^i_t = (x^i_{t+1} - x^i_t, y^i_{t+1} - y^i_t, \psi^i_{t+1} - \psi^i_t, v^i_{t+1} - v^i_t)
\end{align}
We have found that the unconstrained model not only performs worse in terms of the evaluation metrics outlined in Section \ref{sec:eval-metrics}, but also that it sometimes generates physically implausible trajectories, where the cars are driving sideways or even spinning. We speculate that this behavior makes it difficult for the model to use the birdview images as an effective feedback mechanism.

We have also tried using displacements in $x$ and $y$ as the action space, letting the bicycle model fill in the associated orientation and speed, which we refer to as the \emph{displacement model}. Finally, we tried \emph{oriented} versions of both the unconstrained and the displacement models, where the $x$-$y$ coordinates rotate with the ego vehicle. We found the bicycle model to perform much better than any of the alternatives, as shown in Figure \ref{fig:kinematic-ablation}.

\section{Related Work}

The literature on trajectory prediction is vast. Due to space constraints, we narrow our focus to methods that predict distributions over future trajectories with deep generative models, which currently achieve state of the art performance, although even there our coverage is far from complete. For a more comprehensive survey of the field see \cite{rudenko_human_2020}.

The first paper we are aware of to apply CVAEs to the task of motion prediction is \cite{walker_uncertain_2016}, which predicts the movement of pixels in a video of one second into the future from a single frame. The paper is not explicitly concerned with predicting the trajectories of humans navigating through space. One of the most influential papers applying CVAEs to the task of predicting human trajectories is DESIRE \cite{lee_desire_2017}, where like in ITRA the information about the environment is encoded in a birdview image processed by a CNN and a latent random variable is used to generate diverse trajectories. Unlike in ITRA, there is no mechanism for the agents to interact with each other or with the environment past the initial frame.

The most popular alternative to VAEs are generative adversarial networks (GANs) \cite{goodfellow_generative_2014}, which have been been applied to trajectory prediction in papers such as Social GAN \cite{gupta_social_2018}, SoPhie \cite{sadeghian_sophie_2018}, and \cite{zhao_multi-agent_2019}. These approaches differ in how they construct the loss functions and how they encode the information about the past, but in all of them predictions for different agents are generated independently, by decoding a random variable with an RNN.

In PRECOG \cite{rhinehart_precog_2019}, future trajectories of agents are predicted jointly using normalizing flows, another family of deep generative models, to construct a distribution over predicted trajectories. Normalizing flows allow for easy evaluation of likelihoods, which is necessary for model learning, but at the expense of being less flexible than models based on VAEs and GANs. PRECOG models the interactions between agents explicitly, which means it is only applicable to a fixed (or limited) number of agents, while in our model the interactions are mediated through a birdview image, which has a fixed size independent of the number of agents and therefore our model can handle an unbounded number of agents. In PRECOG, prediction is bundled with LiDAR-based detection, which necessitates the use of ad-hoc heuristics to project future LiDAR readings, and precludes it being applied to datasets that do not contain LiDAR readings, such as INTERPRET.

Multiple Futures Prediction \cite{tang_multiple_2019} also predicts the trajectories of different agents jointly, but instead of using continuous latent variables it uses a single discrete variable per agent with a small number of possible values. This allows the model to capture a fixed number of modes in predictions, but is not enough to capture all possible aspects of non-Gaussian variability. It also does not model vehicle orientations. The paper was published before the INTERACTION dataset was released so does not include the model's performance on it and we found the associated code difficult to adapt to this dataset for direct comparison.

TrafficSim \cite{suo_trafficsim_2021}, created independently of and concurrently with our work, proposes a model very similar to ITRA, modelling different vehicles with CVRNNs. It does not utilize differentiable rendering, instead encoding only the static portion of the map as a rasterized birdview image and employing a graph neural network to encode agent interactions. It is not clear to us whether TrafficSim employs any kinematic constraints on the motion of vehicles. The implementation was not released and the paper only reports performance on a proprietary dataset, so we were not able to compare the performance of TrafficSim and ITRA.

\section{Discussion}

The ultimate use cases for multi-agent generative behavioral models like ITRA require additional future work. One problem we found is that when trained on the INTERACTION dataset, ITRA fails to generalize perfectly to significantly different road topologies, sometimes going off-road in non-sensical ways. This issue will most likely be overcome by using additional training data with extensive coverage of road topologies. In recent years, many suitable datasets have been publicly released \cite{caesar_nuscenes_2020, breuer_opendd_2020, krajewski_highd_2018, krajewski_round_2020, houston_one_2020} and training ITRA on them is one of our directions for future work. We note, however, that those datasets come in substantially different formats, particularly regarding the map specification, which requires a significant amount of work to incorporate them into our differentiable simulator.

While the typical use-case for predictive trajectory models is to determine safe areas in path planners, ITRA itself could be used as a planner providing a nominal trajectory for a low-level controller. While this is not necessarily desirable for deployment in autonomous vehicles (AVs), since it imitates both the good and the bad aspects of human behavior, that is precisely what is the goal when building non-playable characters (NPCs) simulating human drivers for training and testing AVs in a virtual environment. We are currently interfacing ITRA with CARLA \cite{dosovitskiy_carla_2017} to control such NPCs, noting that this can be done with any other simulator such as TORCS \cite{wymann_torcs_2014}, LGSVL \cite{rong_lgsvl_2020}, or NVIDIA's Drive Constellation. The main requirement is to provide a low-level controller that can steer the car to follow the trajectory specified by ITRA.

Finally, a drawback of ITRA when used for such applications is that its behavior cannot directly be controlled with high-level commands such as picking the exit to take at a roundabout or yielding to another vehicle. However, since the distribution over trajectories it predicts contains all the behaviors it was exposed to in the dataset, in principle it can be conditioned on achieving certain outcomes, yielding a complete distribution over human-like trajectories that achieve the specified goal. In practice, this is difficult for computational reasons, but can be overcome with techniques such as amortized rejection sampling \cite{naderiparizi_amortized_2019, warrington_coping_2020}, which we are currently working on utilizing in this context.

\section*{APPENDIX}

\subsection{Bicycle Model Equations}

Usually the action space of the kinematic bicycle model consists of acceleration and steering angle and it has two parameters defining the distances from the geometric center to front and rear axes. However, since we do not observe the wheel position and do not know where the axes are, the steering angle can not be determined, so instead we define the steering action to directly control the angle $\beta$ between the vehicle's orientation and its instantaneous velocity.

The continuous time equations of motion for the bicycle model are:
\begin{align}
    \dot{v_t} &= \alpha_t \label{eq:dv} \\
    \dot{x_t} &= v_t \cos (\psi_t + \beta_t) \\
    \dot{y_t} &= v_t \sin (\psi_t + \beta_t) \\
    \dot{\psi_t} &= \frac{v_t}{l_r} \sin (\beta_t) \label{eq:dpsi} .
\end{align}
The action space $a_t = (\alpha_t, \beta_t)$ consists of the acceleration and steering angle applied to the center of the vehicle.
We then use the following discretized version of these equations:
\begin{align}
    v_t &= v_{t-1} + \alpha_t \Delta_t \label{eq:v-update} \\
    x_t &= x_{t-1} + v_t \cos (\psi_{t-1} + \beta_t) \Delta_t \\
    y_t &= y_{t-1} + v_t \sin (\psi_{t-1} + \beta_t) \Delta_t \\
    \psi_t &= \psi_{t-1} + \frac{v_t}{l_r} \sin (\beta_t) \Delta_t \label{eq:psi-update} .
\end{align}

\subsection{Fitting the Bicycle Model} \label{sec:bicycle-fitting}

The dataset contains a sequence of states for each vehicle, but not directly the actions of the bicycle model. To recover the actions, we find the values that match the displacements recorded in the dataset. Specifically, we compute the acceleration and steering according to the following equations, for $t>1$:
\begin{align}
    \alpha_t &= \frac{1}{\Delta_t} (\frac{1}{\Delta_t} \sqrt{(x_t^{g.t.} - x_{t-1})^2 + (y_t^{g.t.} - y_{t-1})^2} - v_{t-1}) \label{eq:a-fit} \\
    \beta_t &= \mathop{\atantwo}(\frac{1}{\Delta_t} (y_t^{g.t.} - y_{t-1}), \frac{1}{\Delta_t} (x_t^{g.t.} - x_{t-1})) - \psi_{t-1} \label{eq:beta-fit} ,
\end{align}
where $g.t.$ superscript indicates that the values are the ground truth from the dataset, otherwise they are obtained from the previous time step.

The bicycle model fits the observed trajectories closely but not exactly.
We define the loss measuring how well the bicycle model fits a given trajectory as
\begin{align}
    L_{bicycle-fit} = \max_{t \in \{2, \dots, T\}} 2 (1 - \cos(\psi_t - \psi_t^{g.t.})),
\end{align}
which is chosen to symmetrically increase with the discrepancy angle in both directions, also to be $2\pi$-periodic.

\begin{figure}[t]
    \centering
    \includegraphics[width=\columnwidth]{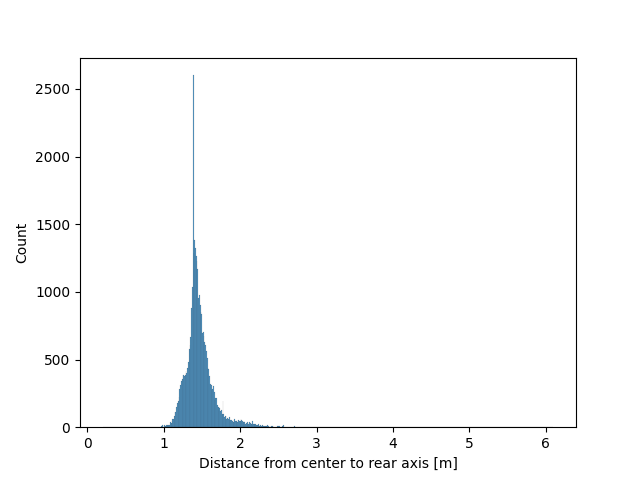}
    \includegraphics[width=\columnwidth]{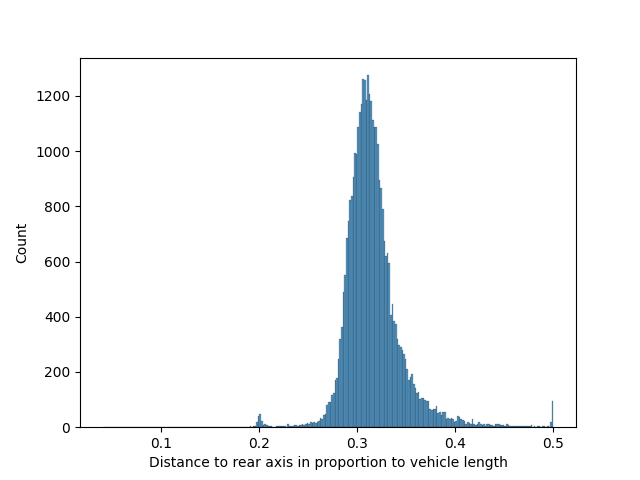}
    \caption{The values of distance from vehicle center to its rear axis in proportion to vehicle length, found by grid search fitting the bicycle model. Note that for about 100 vehicles the value of $L^i/2$ fits best, which is the end of the grid search range and physically plausible. Possibly these cars are going straight for the whole track and the distance to rear axis can not be determined, but its larger values reduce the overfitting of the bicycle model to the noise in annotations.}
    \label{fig:lr-hist}
\end{figure}

The one missing parameter is $l_r$, which is not available in the dataset. To find the corresponding values, we perform a grid search on $l_r^i$ in order to minimize $L_{bicycle-fit}^i$ separately for each vehicle $i$. We search in $1$ cm increments up to half the vehicle length. The histogram of the resulting values of $l_r$ is shown in Figure \ref{fig:lr-hist}. Using this procedure, we achieve a near-perfect fit of the trajectories recorded in the dataset.

\section*{Acknowledgements}

This research was primarily funded by the Mitacs Accelerate grant IT16342.
We acknowledge the support of the Natural Sciences and Engineering Research Council of Canada (NSERC), the Canada CIFAR AI Chairs Program, and the Intel Parallel Computing Centers program. This material is based upon work supported by the United States Air Force Research Laboratory (AFRL) under the Defense Advanced Research Projects Agency (DARPA) Data Driven Discovery Models (D3M) program (Contract No. FA8750-19-2-0222) and Learning with Less Labels (LwLL) program (Contract No.FA8750-19-C-0515). Additional support was provided by UBC's Composites Research Network (CRN), Data Science Institute (DSI) and Support for Teams to Advance Interdisciplinary Research (STAIR) Grants. This research was enabled in part by technical support and computational resources provided by WestGrid (https://www.westgrid.ca/), Compute Canada (www.computecanada.ca), and Weights and Biases (wandb.ai).

We thank Wilder Lavington and Michiel van de Panne for the helpful discussions.

\addtolength{\textheight}{-7cm}   




\bibliographystyle{IEEEtran}
\bibliography{iai-refs}

\clearpage

\begin{figure*}[t]
    \centering
    \begin{subfigure}[b]{0.31\textwidth}
        \centering
        \includegraphics[width=\textwidth, frame]{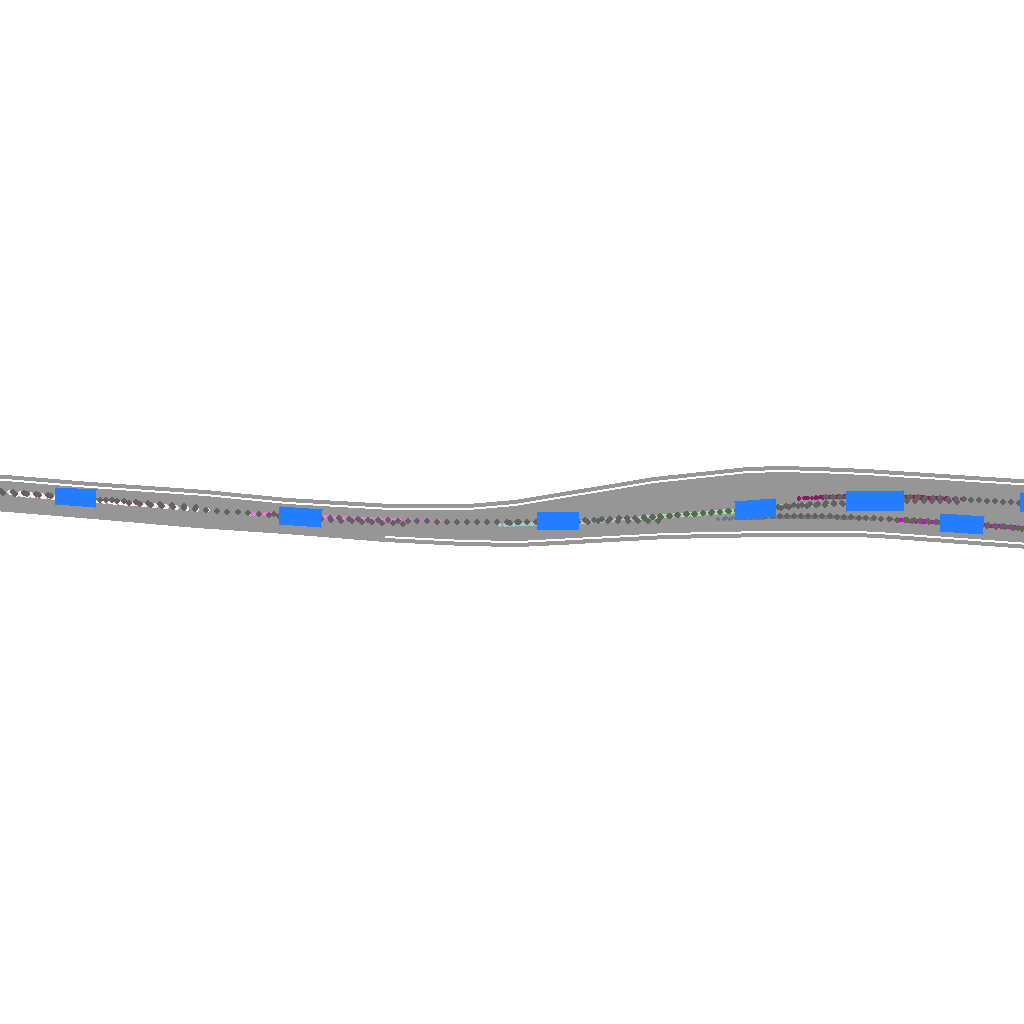}
        \caption{DR\_DEU\_Merging\_MT}
    \end{subfigure}
    \hfill
    \begin{subfigure}[b]{0.31\textwidth}
        \centering
        \includegraphics[width=\textwidth,frame]{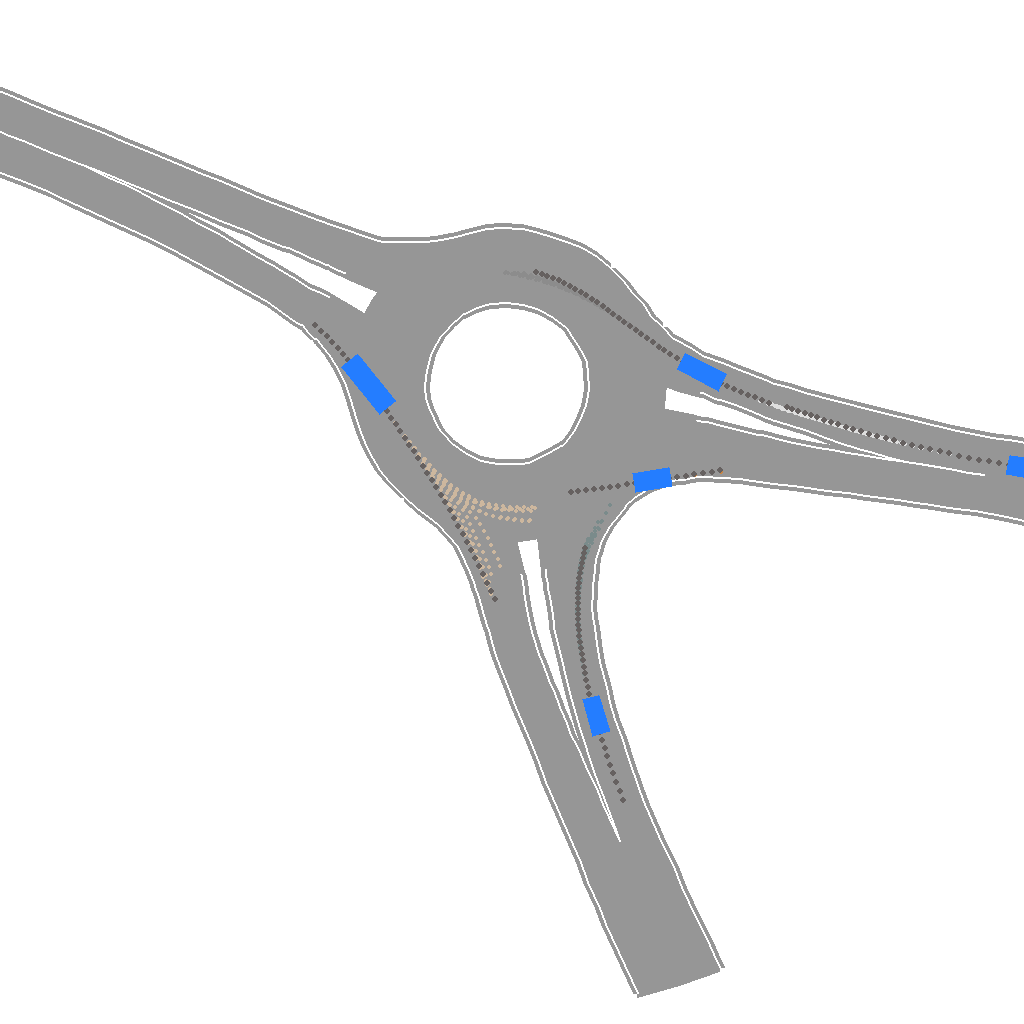}
        \caption{DR\_DEU\_Roundabout\_OF}
    \end{subfigure}
    \hfill
    \begin{subfigure}[b]{0.31\textwidth}
        \centering
        \includegraphics[width=\textwidth, frame]{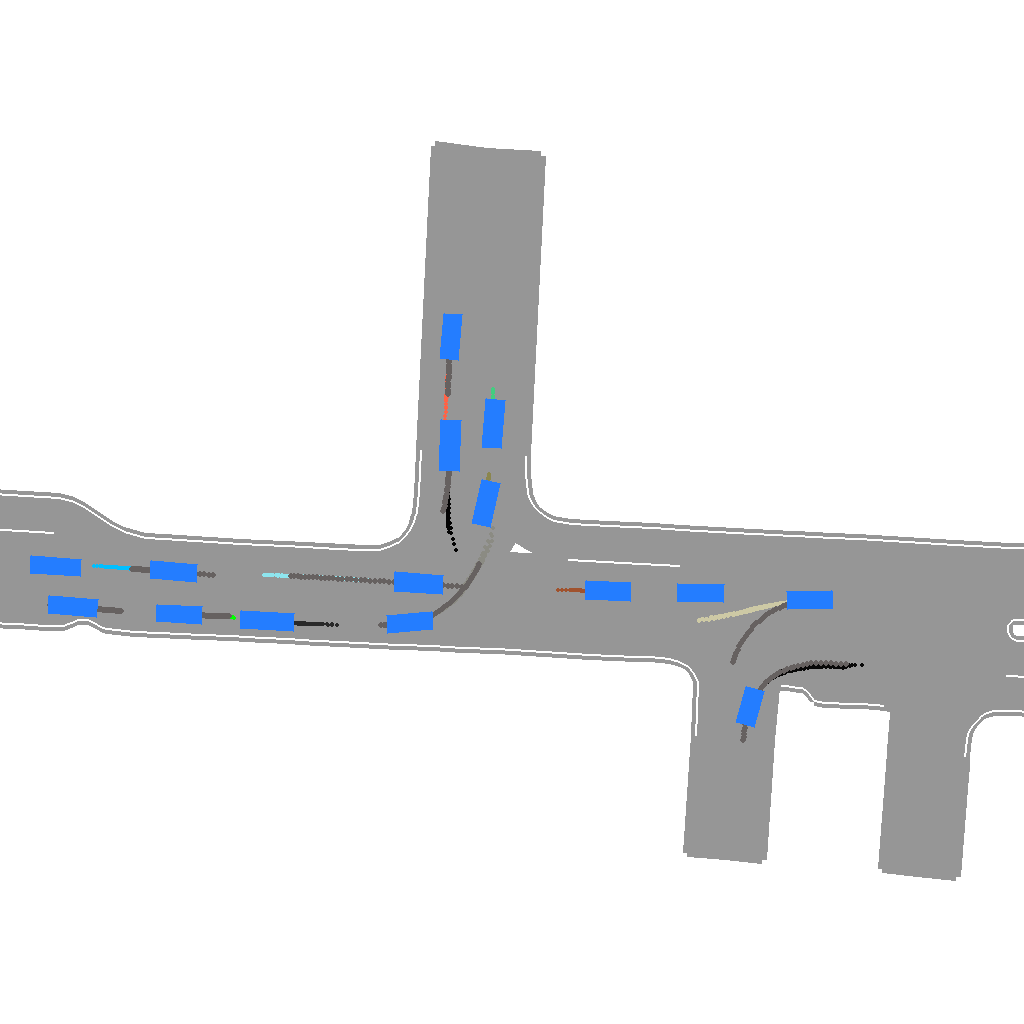}
        \caption{DR\_USA\_Intersection\_EP0}
    \end{subfigure}
    \hfill
    \begin{subfigure}[b]{0.31\textwidth}
        \centering
        \includegraphics[width=\textwidth, frame]{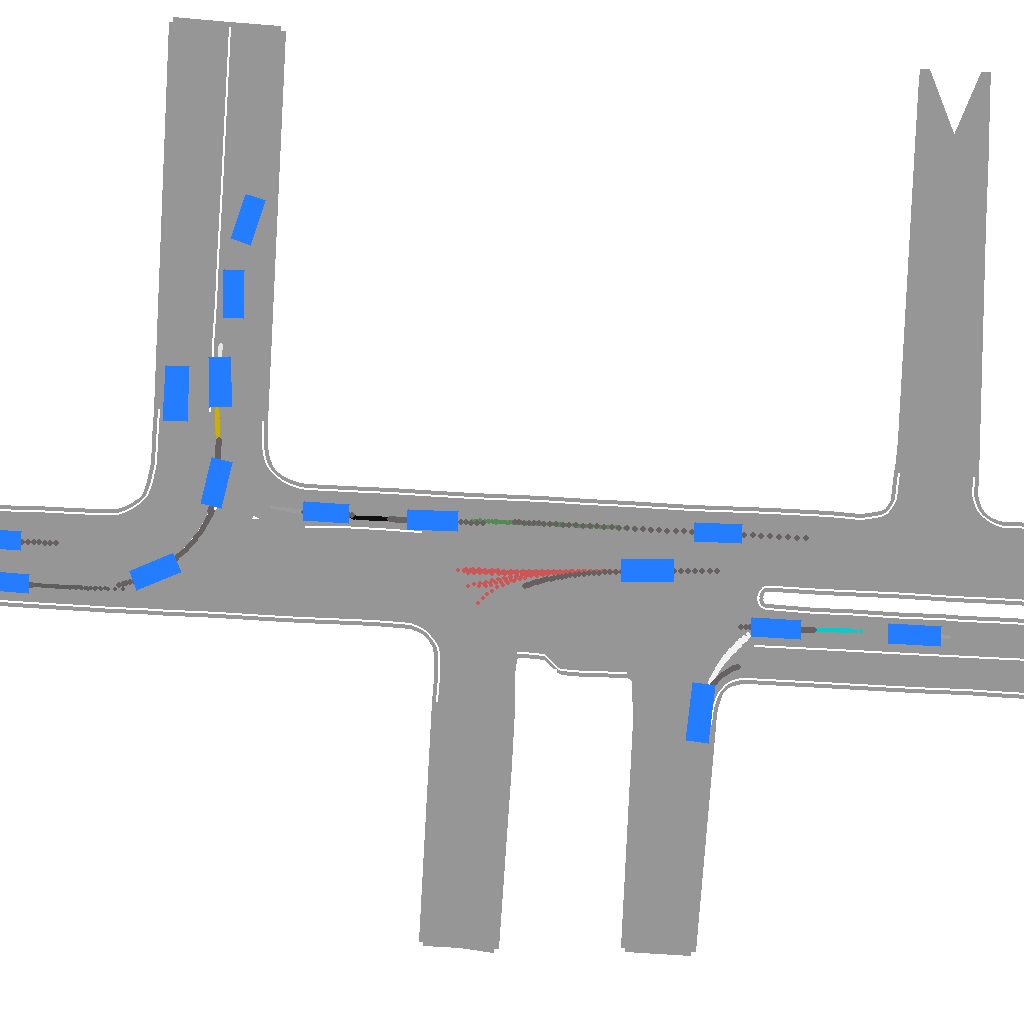}
        \caption{DR\_USA\_Intersection\_EP1}
    \end{subfigure}
    \hfill
    \begin{subfigure}[b]{0.31\textwidth}
        \centering
        \includegraphics[width=\textwidth, frame]{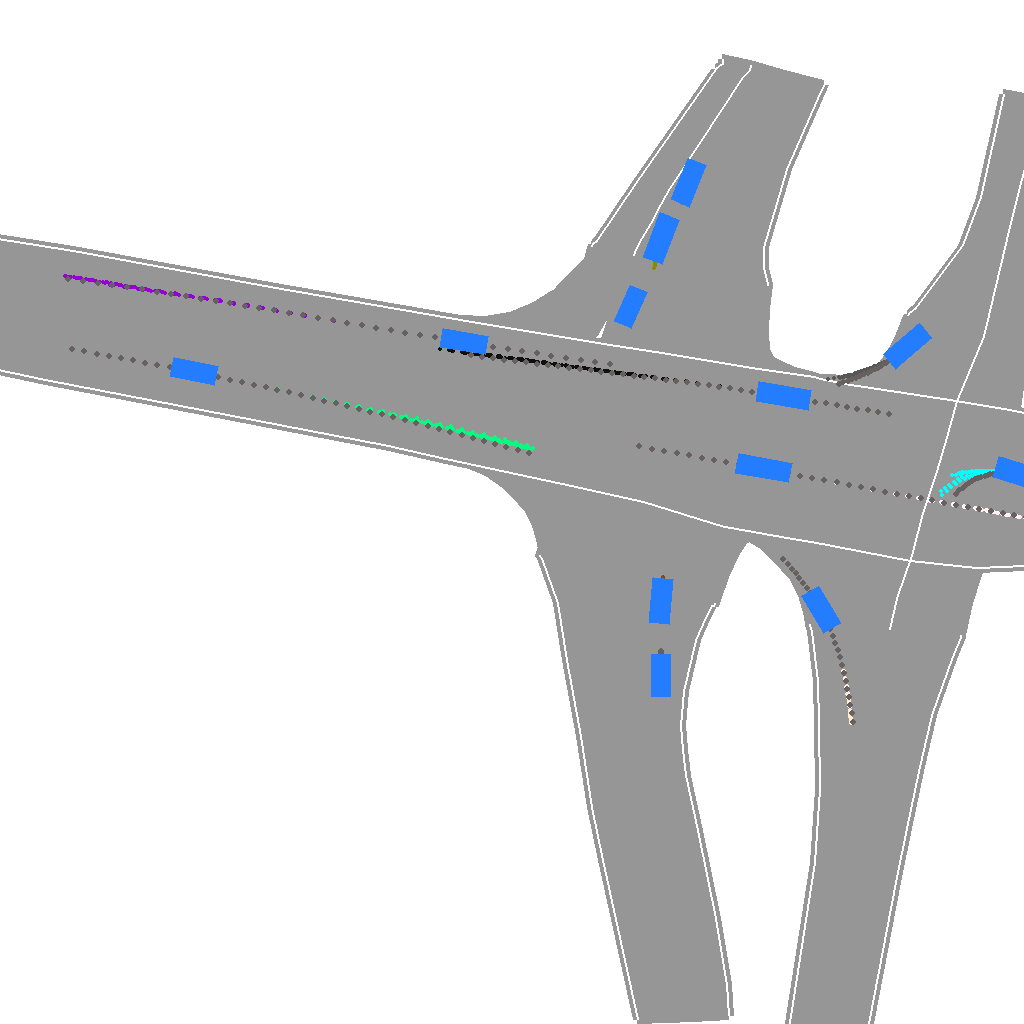}
        \caption{DR\_USA\_Intersection\_GL}
    \end{subfigure}
    \hfill
    \begin{subfigure}[b]{0.31\textwidth}
        \centering
        \includegraphics[width=\textwidth, frame]{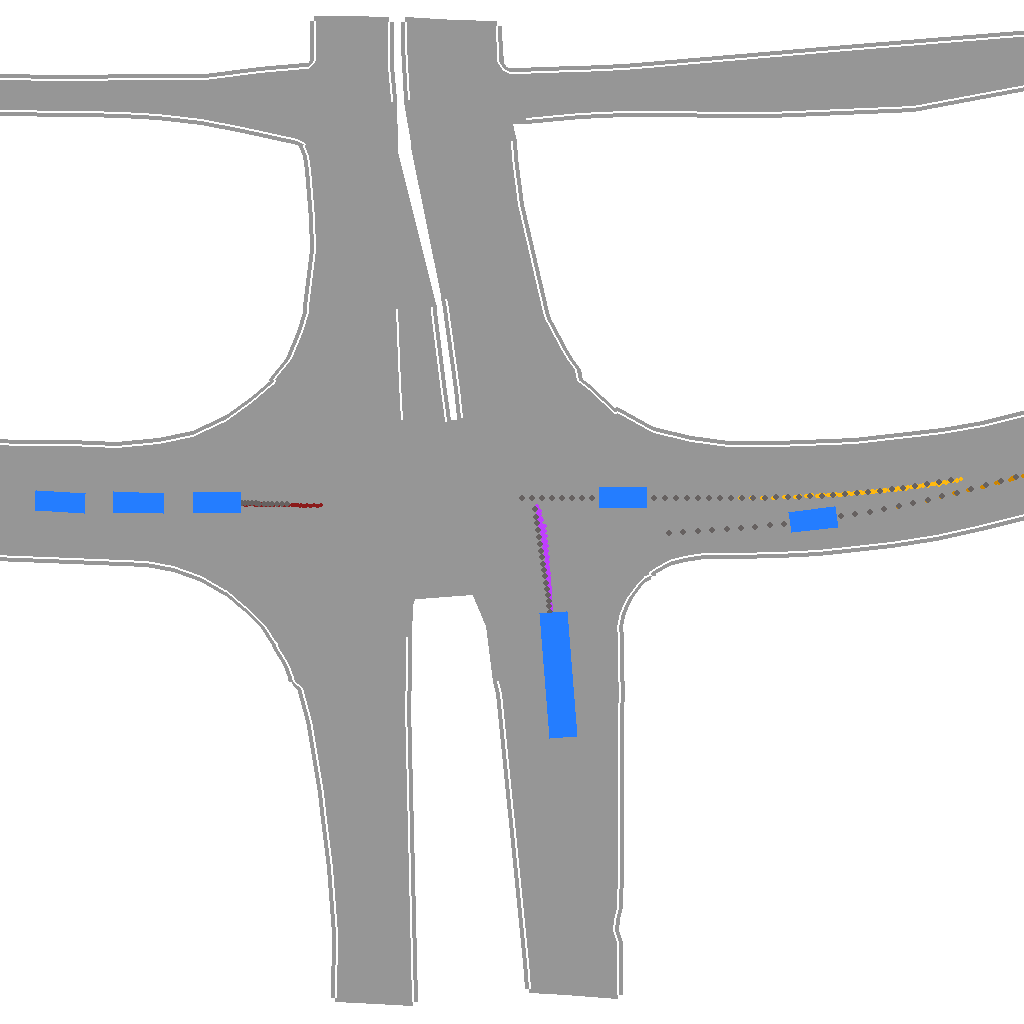}
        \caption{DR\_USA\_Intersection\_MA}
    \end{subfigure}
    \hfill
    \begin{subfigure}[b]{0.31\textwidth}
        \centering
        \includegraphics[width=\textwidth, frame]{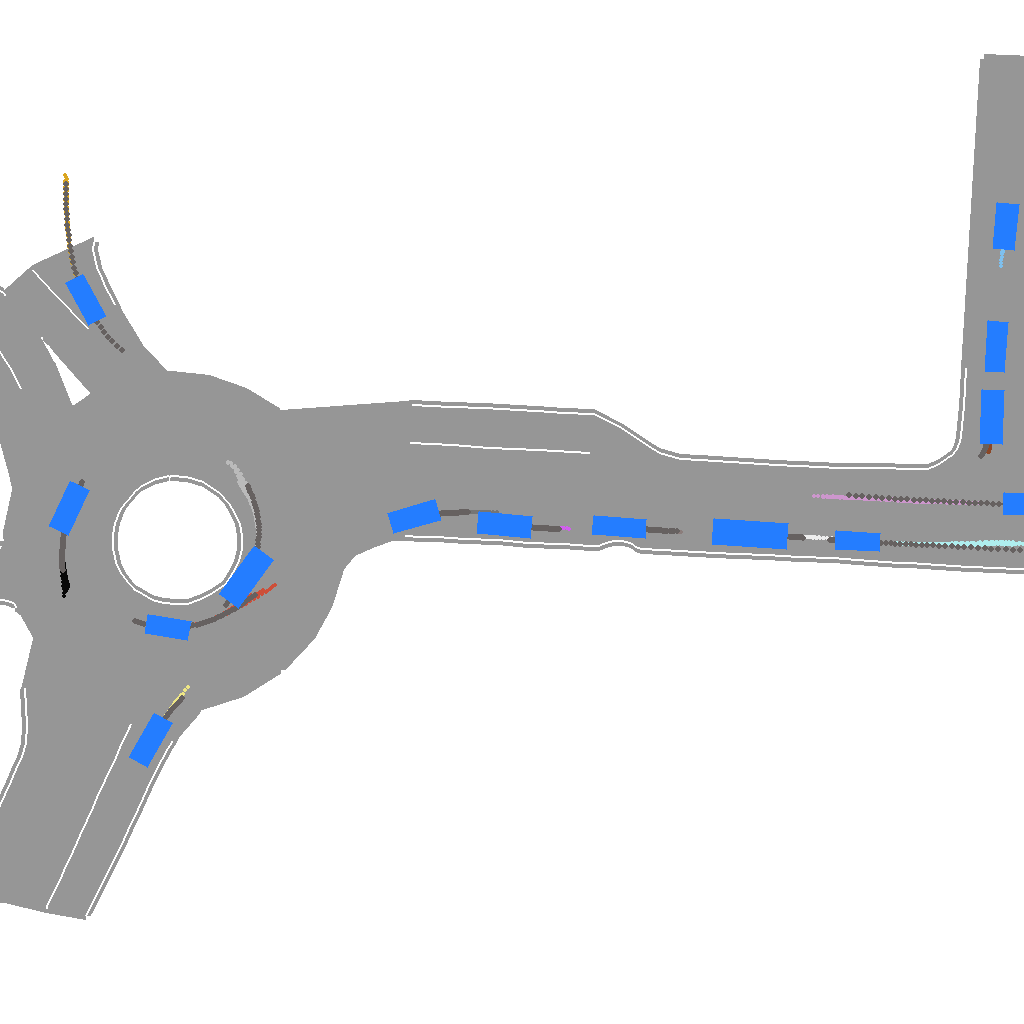}
        \caption{DR\_USA\_Roundabout\_EP}
    \end{subfigure}
    \hfill
    \begin{subfigure}[b]{0.31\textwidth}
        \centering
        \includegraphics[width=\textwidth, frame]{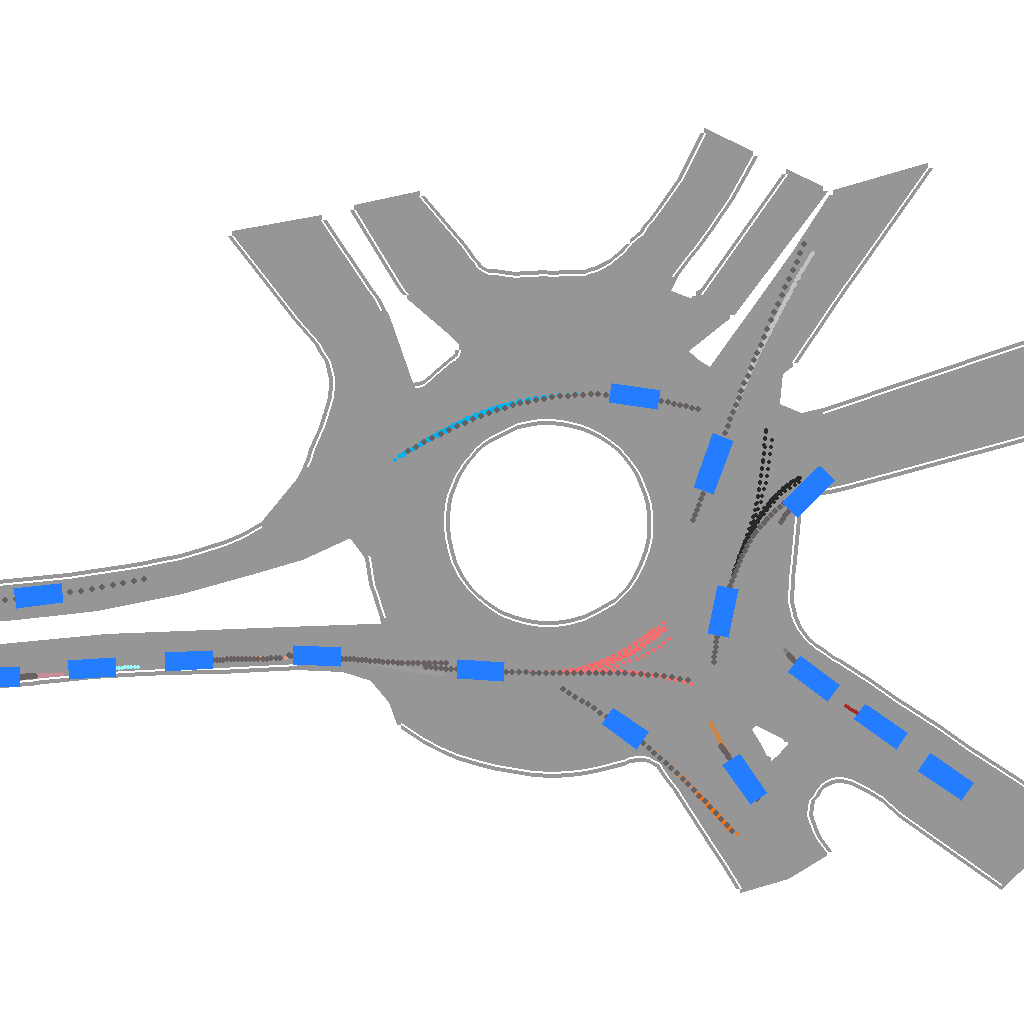}
        \caption{DR\_USA\_Roundabout\_FT}
    \end{subfigure}
    \hfill
    \begin{subfigure}[b]{0.31\textwidth}
        \centering
        \includegraphics[width=\textwidth, frame]{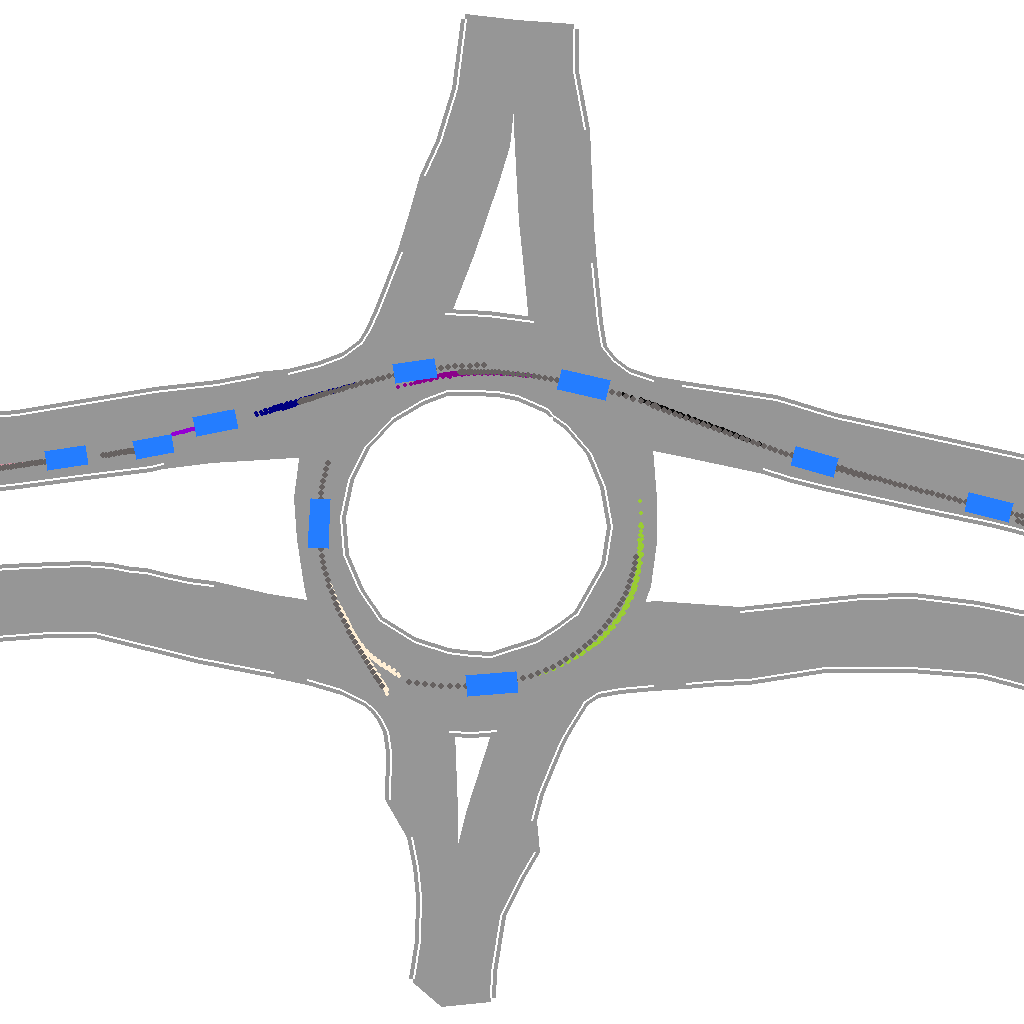}
        \caption{DR\_USA\_Roundabout\_SR}
    \end{subfigure}
    \hfill
    \begin{subfigure}[b]{0.31\textwidth}
        \centering
        \includegraphics[width=\textwidth, frame]{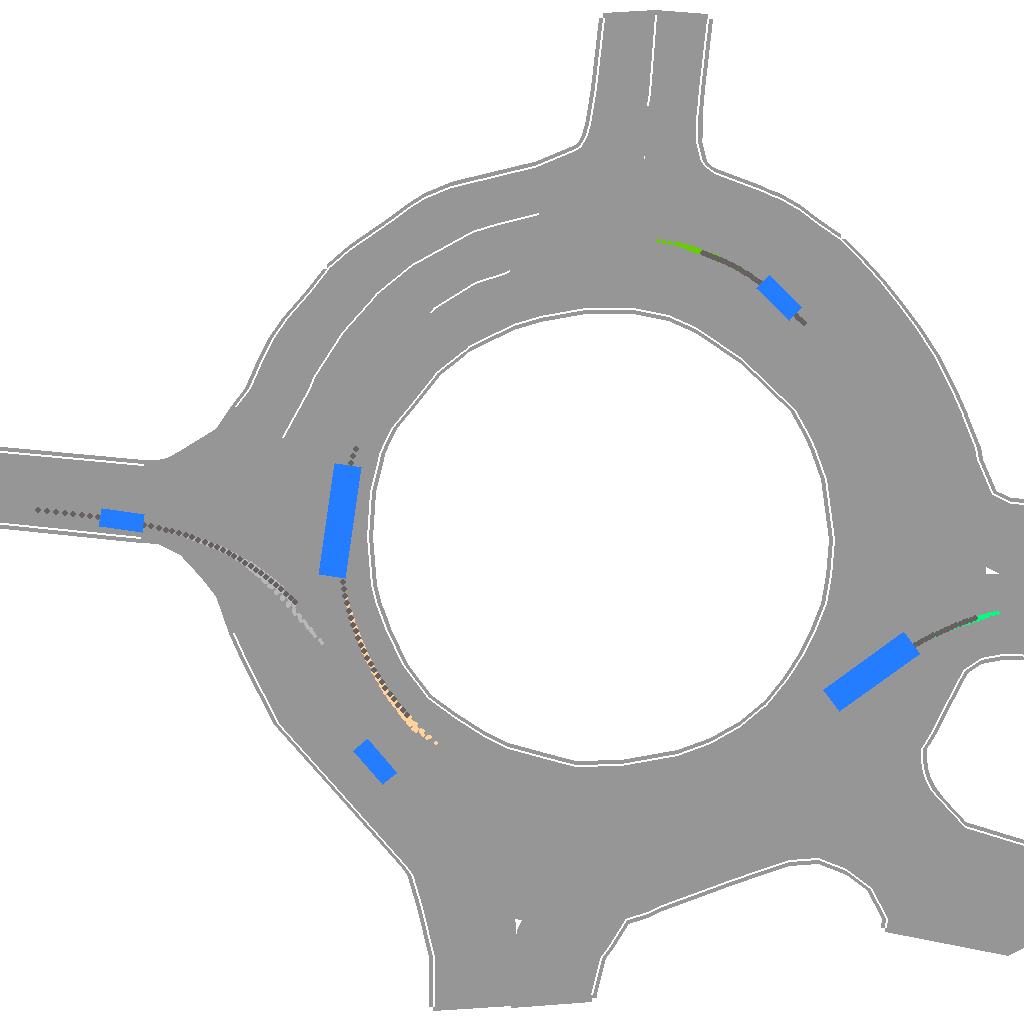}
        \caption{DR\_CHN\_Roundabout\_LN}
    \end{subfigure}
    \hspace{1em}
    \begin{subfigure}[b]{0.31\textwidth}
        \centering
        \includegraphics[width=\textwidth, frame]{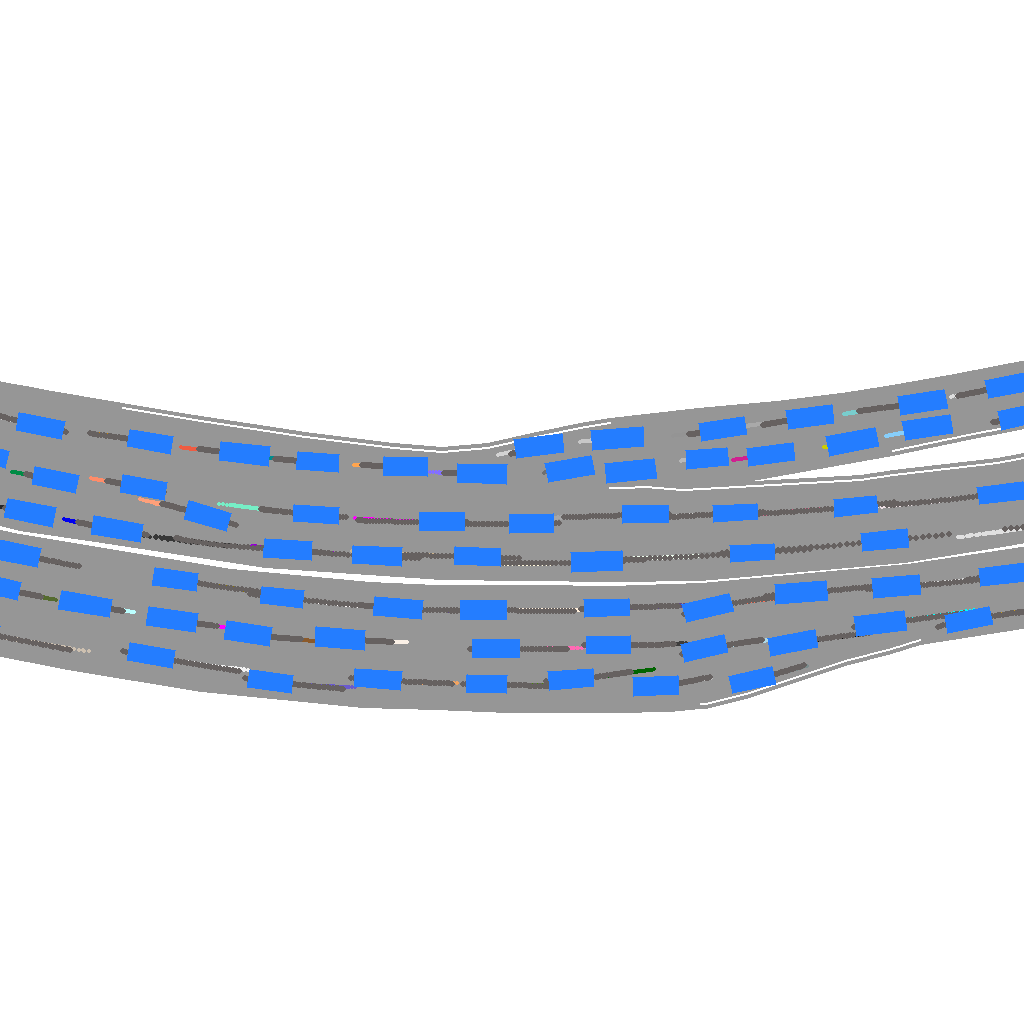}
        \caption{DR\_CHN\_Merging\_ZS}
    \end{subfigure}
    \caption{Example predictions of ITRA for each available map in the INTERACTION dataset \cite{zhan_interaction_2019}. Each car is displayed with the ground truth trajectory (in dark grey) and 10 different future predicted trajectories.}
    \label{fig:birdview_examples}
\end{figure*}

\end{document}